\documentclass[acmsmall]{acmart}

\AtBeginDocument{%
  \providecommand\BibTeX{{%
    \normalfont B\kern-0.5em{\scshape i\kern-0.25em b}\kern-0.8em\TeX}}}

\setcopyright{acmcopyright}
\copyrightyear{2019}
\acmYear{2019}
\acmDOI{xx.xxxx/xxxxxxx.xxxxxxx}

\usepackage{booktabs} 
\usepackage{epstopdf}
\usepackage{url}
\usepackage{textcomp}
\usepackage{algorithmic}
\usepackage{algorithm}
\usepackage{bm}
\usepackage{amssymb}
\usepackage{pdfrender}
\usepackage{subcaption}
\usepackage{multirow}
\newcommand*{\boldcheckmark}{%
  \textpdfrender{
    TextRenderingMode=FillStroke,
    LineWidth=.5pt, 
  }{\checkmark}%
}

\acmJournal{TECS}
\acmVolume{00}
\acmNumber{0}
\acmArticle{000}
\acmMonth{0}

\begin{document}

\title[Network of Neural Networks]{Memory- and Communication-Aware Model Compression for Distributed Deep Learning Inference on IoT}

\titlenote{This preprint is for personal use only. The official article will appear as part of the ESWEEK-TECS special issue and will be presented in the International Conference on Hardware/Software Codesign and System Synthesis (CODES+ISSS), 2019}

\author{Kartikeya Bhardwaj}
\email{kbhardwa@cmu.edu}
\orcid{0000-0002-8115-4276}
\author{Chingyi Lin}
\email{chingyil@andrew.cmu.edu}
\author{Anderson Sartor}
\email{asartor@cmu.edu}
\author{Radu Marculescu}
\email{radum@cmu.edu}
\affiliation{%
  \institution{\\Carnegie Mellon University}
  \streetaddress{5000 Forbes Avenue}
  \city{Pittsburgh}
  \state{PA}
  \country{USA}
  \postcode{15213}
}

\renewcommand{\shortauthors}{Bhardwaj, et al.}

\begin{abstract}
Model compression has emerged as an important area of research for deploying deep learning models on Internet-of-Things (IoT). However, for extremely memory-constrained scenarios, even the compressed models cannot fit within the memory of a single device and, as a result, must be distributed across multiple devices. This leads to a \textit{distributed inference} paradigm in which memory and communication costs represent a major bottleneck. Yet, existing model compression techniques are not communication-aware. Therefore, we propose \textit{Network of Neural Networks} (NoNN),  a new distributed IoT learning paradigm that compresses a large pretrained `teacher' deep network into several disjoint and highly-compressed `student' modules, without loss of accuracy. Moreover, we propose a network science-based knowledge partitioning algorithm for the teacher model, and then train individual students on the resulting disjoint partitions. Extensive experimentation on five image classification datasets, for user-defined memory/performance budgets, show that NoNN achieves higher accuracy than several baselines and similar accuracy as the teacher model, while using minimal communication among students. Finally, as a case study, we deploy the proposed model for CIFAR-10 dataset on edge devices and demonstrate significant improvements in memory footprint (up to $24\times$), performance (up to $12\times$), and energy per node (up to $14\times$) compared to the large teacher model. We further show that for distributed inference on multiple edge devices, our proposed NoNN model results in up to $33\times$ reduction in total latency w.r.t. a state-of-the-art model compression baseline.
\end{abstract}



\maketitle

\section{Introduction}
Even though deep learning has gained significant importance, it is challenging to implement these models on resource-constrained devices. Such devices are commonly used in the ever-growing \textit{Internet-of-Things} (IoT) domain.  For instance, microcontrollers such as Arm Cortex-M which is used in many IoT applications such as Smart Healthcare, Keyword Spotting, \textit{etc.}, has only 500KB available memory~\cite{kws}. For such devices, prominent deep networks like AlexNet/Resnets that use up to $60$M parameters are unsuitable. Therefore, there is a fundamental need for highly-efficient compressed deep learning models to enable 
\textit{fast and computationally inexpensive inference} on the resource-constrained IoT-devices.

\begin{figure}[tb]
\centering
\includegraphics[width=0.8\textwidth]{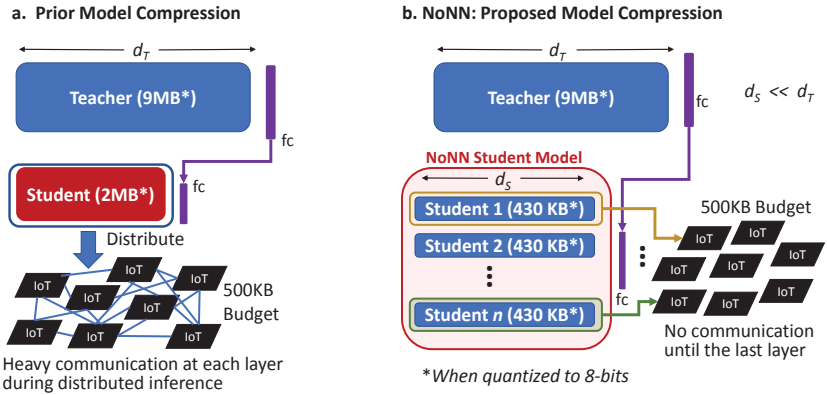}
\caption{(a) Prior art: Distributing large student models that do not fit on a single memory-limited IoT device leads to heavy communication at each layer. For instance, a student model that requires 2MB total memory cannot fit within an IoT-device which has only 500KB memory budget. (b) Proposed NoNN results in several disjoint students that can fit on individual IoT-devices: No communication until the very final layer (since NoNN students that require 430KB memory can fit within the 500KB budget).}
\label{kdFig}
\end{figure}

To this end, several deep learning model compression techniques exist in the literature. The most common approaches are pruning~\cite{hanNIPS2015, prune3}, quantization~\cite{quant2, floatFix}, and Knowledge Distillation (KD)~\cite{deepShallow1, hintonKD} and its variants such as Attention Transfer (AT)~\cite{atkd}. Since deep models can learn a large number of redundant or useless weights/channels, pruning aims to remove such parameters without sacrificing accuracy~\cite{hanNIPS2015, prune3}. Moreover, quantization reduces the number of bits required to represent weights and/or activations in deep networks~\cite{quant2, floatFix}. Also, KD-based approaches rely on teacher-student training, where a teacher model is a large deep network we want to compress. KD trains a significantly smaller student model with far fewer layers/width, to mimic the teacher network (see Fig.~\ref{kdFig}a). Further, most prior art on model compression refers to pruning and quantization. However, since KD can lead to significantly compressed models, we assume that KD is also a model compression technique throughout this paper. Of note, the present work considers model compression and distributed inference for Convolutional Neural Networks (CNNs) used in image classification problems. However, distributed inference for other deep learning models such as Recurrent Neural Networks (RNNs) for speech/natural language applications can also be explored in a future work.

The existing approaches mentioned above \textit{cannot} be used for \textit{extremely memory-constrained} IoT scenarios (\textit{e.g.}, microcontrollers with a \textit{total} memory of 500KB~\cite{kws,stMicro}). Yet, a lot of smart home/cities applications can have several such connected but resource-limited devices. To achieve higher accuracy, pruning- or KD-based student models often grow in size due to which such models cannot fit on an individual IoT-device and must be distributed across multiple devices; this distribution of computation on multiple devices generates significant  overhead in communication. For example, Fig.~\ref{kdFig}a illustrates KD with a teacher model containing around 9 million parameters, and a large student model containing about 2 million parameters. Also, suppose that we are trying to deploy the student model (which will require $2$MB memory after quantization) on IoT-devices with only 500KB memory budget. Distributing this student model on multiple such IoT-devices will incur heavy communication at each of its intermediate layers. Consequently, in addition to the computation cost, a major (and largely ignored) impediment for widespread deployment of deep learning on IoT is this \textit{communication} cost during distributed inference, which occurs due to extremely limited memory in IoT environments. 

Therefore, we need to answer the following \textbf{key question}: 
\textit{How can we compress a given deep learning model into multiple separate modules that: (i) fit within the memory and performance budgets per device, (ii) minimize the communication latency when distributed across a network of edge devices, and (iii) achieve high accuracy?}

To address this question, in this paper, we propose \textit{Network of Neural Networks} (NoNN), a paradigm to derive low memory-, computation-, as well as communication-cost student architectures from a single (powerful) but much larger teacher model. As shown in Fig.~\ref{kdFig}b, a NoNN consists of multiple, \textit{memory-limited} student networks which individually learn only a specific part of teacher's function. 
Indeed, training an individual student to mimic only  a part of teacher's function is a significantly simpler problem than mimicking its entire function. Since this highly-parallel architecture of NoNN allows us to effectively increase the overall model-size, we obtain higher accuracy without significantly increasing the communication costs among the disjoint students. These individual students can then be deployed on separate resource-constrained IoT-devices to perform the distributed inference (see Fig.~\ref{kdFig}b). This can ultimately help preserve user's privacy by conducting inference on a network of edge devices instead of sending data to the cloud.

Finally, to distribute knowledge from a teacher to multiple students, we propose a new approach based on network science~\cite{networksci, newmanComm}. Specifically, as our objective is to obtain individual student models below certain memory-constraints, network science allows for significant flexibility in terms of how the knowledge from the teacher is distributed across multiple students. For instance, with our proposed network-theoretic approach, we uniformly distribute knowledge from the teacher across the students to achieve high accuracy.

Overall, we make the following \textbf{key contributions}:
\begin{enumerate}
\item Since there has been limited research on performing efficient distibuted inference on highly memory-constrained IoT devices, we propose a new NoNN paradigm to compress a large teacher into a set of independent and highly-parallel student networks. In prior works, student models that cannot fit on a single device can lead to significant communication at all layers, whereas NoNN communicates only at the last layer. This makes NoNN an ideal candidate for distributed deep learning. To the best of our knowledge, we are the \textit{first} to introduce a communication-aware model compression for distributed inference on IoT-devices.
\item We are also the first to formulate the model compression problem from a network science angle. By building a network of filter activation patterns, we exploit principles from network science such as community detection~\cite{newmanComm} to split teacher's knowledge into multiple disjoint partitions which, in turn, results in disjoint and compressed student modules. 
\item Extensive experiments on five well-known image classification tasks demonstrate that NoNN achieves significantly lower memory ($2.5\times$-$24\times$ reduction w.r.t. teacher) and computation ($2\times$-$15\times$ fewer FLOPS w.r.t. teacher), with minimal communication costs and similar accuracy as the teacher model. Also, NoNN achieves higher classification accuracy than prior art~\cite{hintonKD, atkd, splitnet}.
\item We further deploy the proposed models for CIFAR-10 dataset on Raspberry Pi, and more powerful Odroid devices to evaluate several scenarios with homogeneous and heterogeneous devices. We demonstrate $6.22\times$-$12.22\times$ gain in performance and $12.99\times$-$14.36\times$ gain in per node energy w.r.t. teacher. Moreover, for distributed inference on multiple edge devices, NoNN results in up to $33\times$ speedup in total latency w.r.t. a prior model compression baseline~\cite{atkd}. We also show that the proposed approach is robust and achieves high accuracy even with a reduced number of devices.
\end{enumerate}

The rest of the paper is organized as follows: Section 2 reviews relevant literature and background. The impact of splitting a deep network horizontally on distributed inference performance is discussed in Section 3. Section 4 describes our proposed approach, while Section 5 presents our detailed experimental results. Section 6 describes a case study on hardware deployment of NoNN. The paper is finally concluded in Section 7 with remarks on future work.

\section{Related Work and Background}
We first discuss prior art on model compression and distributed inference. The relevant background on Knowledge Distillation (KD), and network science is discussed later in this section.

\subsection{Model Compression/Distributed Inference}
Table~\ref{rel} summarizes the key differences between the present work and the existing literature from model compression and distributed inference.
\begin{table}[tb]
\centering
\caption{Comparison to Prior Art. NoNN is the most complete approach as it considers communication costs while performing model compression for distributed inference on IoT.}
\scalebox{0.74}{
\label{rel}
\begin{tabular}{l|c|c|c|c} 
\hline
Area& Model& Communication-& Distributed& Complements\\
& Compression& Aware& Inference& our work\\
\hline
Quantization~\cite{floatFix, quant2}& $\boldcheckmark$ & $\bm{\times}$& $\bm{\times}$& $\boldcheckmark$\\
Pruning~\cite{prune2, prune3, sqn, deepComp}& $\boldcheckmark$& $\bm{\times}$& $\bm{\times}$& $\boldcheckmark$\\
Separable convolutions~\cite{mobilenetV2, shufflenets, channelnets}& $\boldcheckmark$& $\bm{\times}$& $\bm{\times}$& $\boldcheckmark$\\
KD~\cite{deepShallow1, hintonKD, atkd}& $\boldcheckmark$& $\bm{\times}$& $\bm{\times}$& $\bm{\times}$\\
SplitNet~\cite{splitnet} & $\bm{\times}$ & $\boldcheckmark$ & $\boldcheckmark$ & $\bm{\times}$\\
MoDNN, DeepThings~\cite{modnn,deepthings}& $\bm{\times}$ & $\boldcheckmark$ & $\boldcheckmark$ & $\boldcheckmark$\\ \hline
\textbf{Proposed NoNN}& $\boldcheckmark$ & $\boldcheckmark$ & $\boldcheckmark$ & $\boldcheckmark$  \\ \hline
\end{tabular}
}
\end{table}
Specifically, existing approaches for model compression such as quantization~\cite{floatFix, quant2}, pruning~\cite{prune2, prune3, deepComp}, KD~\cite{deepShallow1, hintonKD, atkd}, and separable convolutions~\cite{mobilenetV2, shufflenets, channelnets} are \textit{not} communication-aware and do \textit{not} address the distributed inference problem. For instance, although pruning is effective at reducing parameters/FLOPS, it does \textit{not} result in highly-parallel model architectures. Clearly, pruned models that cannot fit on a single IoT-device will need to be distributed across multiple devices which will incur heavy communication costs (see Section 3 for details). 

On the other hand, existing distributed inference techniques such as SplitNet~\cite{splitnet}, MoDNN~\cite{modnn}, and DeepThings~\cite{deepthings} do not exploit \textit{ideas} from model compression (\textit{e.g.}, pruning/distillation, or memory-constraints). Specifically, SplitNet~\cite{splitnet} splits the network into disjoint parts during training without any constraints on individual network partitions. This unconstrained splitting can result in large partitions which may not conform to memory-budgets of individual IoT devices. Hence, 
taking specific memory budgets into account is more effective (and necessary) in order to meet hard memory-constraints of IoT-devices; such constraints are not considered by SplitNet~\cite{splitnet}. 
Moreover, by partitioning the feature activation maps of convolutions, MoDNN~\cite{modnn} and DeepThings~\cite{deepthings} successfully reduce the FLOPS and the feature activation map memory during distributed inference. However, both of these approaches assume that the mobile device is big enough to fit the entire deep learning model, which is a strong assumption   and, hence, they do not address the memory due to model-weights. In contrast, we take the distributed inference one step further to an even more constrained IoT environment (\textit{e.g.}, 500KB memory) where the physical IoT-devices cannot fit a single model. In such cases, the model itself must be split in non-intuitive ways which can result in heavy communication cost at every layer. Therefore, in this paper, our proposed NoNN compresses the model (to reduce the memory and computation costs such that individual modules fit on the devices) while accounting for communication among the student nodes. 

Note that, most of the prior techniques above are complementary to our work so they can be used synergistically on top of our approach. For instance, another recent work called MeDNN~\cite{mednn} extends MoDNN with a pruning technique to reduce the per-node computation. Since pruning can still be performed on top of our proposed NoNN and because both MeDNN and MoDNN partition the feature maps (and not the weights), our work is complementary to MeDNN~\cite{mednn}. We also show that we can achieve higher accuracy than several KD baselines and SplitNet in our experiments, \textit{i.e.}, the only two techniques in Table~\ref{rel} that do not complement our work.

\subsection{Knowledge Distillation Background}
\begin{figure}[]
\centering
\includegraphics[width=0.7\textwidth]{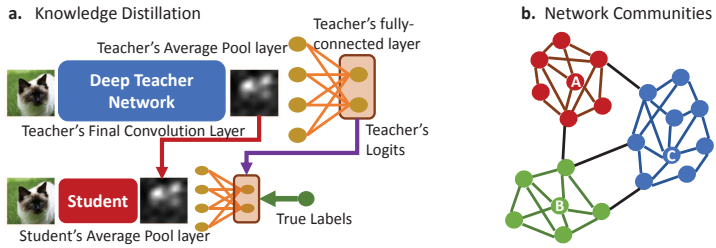}
	\caption{(a) Knowledge Distillation (KD) is based on a significantly smaller student model trained to mimic a large teacher network. (b) Network communities and hub nodes. In Section 4, we will explain how teacher's activations from (a) can be used to create a network similar to that shown in (b).}
\label{kdb}
\end{figure}

KD, as shown in Fig.~\ref{kdb}a, consists of two deep networks: (\textit{i}) \textit{Teacher} is the large deep network which we want to compress, and (\textit{ii}) \textit{Student} is a significantly smaller neural network which is trained to mimic the output of the teacher network. To mimic the teacher, the simplest way is to directly train a student model on teacher's \textit{logits} (see Fig.~\ref{kdb}a) -- the unnormalized outputs of the teacher model before the softmax -- instead of training on true labels from the dataset~\cite{deepShallow1}. The basic idea here is that while training the student model, using logits instead of direct classification probabilities can transfer more information about the teacher to the student. Hinton \textit{et al.}~\cite{hintonKD} argued that we must not only train the student model on the correct predictions made by the teacher, but must also quantify how wrong the teacher was about the incorrect classes. Hence, Hinton \textit{et al.} introduced KD~\cite{hintonKD} which uses both the \textit{hard-label loss} (based on true labels from the dataset), as well as the \textit{soft-label loss} (based on logits) to train a significantly smaller student model.

Mathematically, let $l_T$ and $l_S$ be the logits of teacher and student respectively, and $\tau$ is a temperature parameter (see~\cite{hintonKD}), $y$ be the true labels, and $P_T^\tau$ and $P_S^\tau$ respectively denote the softmax over relaxed logits $l_T/\tau$ and $l_S/\tau$. Then, the KD loss ($\mathcal{L}^{kd}$) is given by:
\begin{equation}
\mathcal{L}^{kd}(\theta_S)=(1-\alpha)\mathbb{H}(y,P_S)+\alpha \mathbb{H}(P_T^\tau,P_S^\tau)
\label{kdloss1}
\end{equation}
where, $\mathbb{H}$ is the standard cross-entropy loss, $\theta_S$ denotes the parameters of the student network, and $\alpha$ controls the weight of hard-label loss \textit{vs.} soft-label loss. The temperature parameter in the second term of Eq.~\eqref{kdloss1} improves knowledge transfer from the teacher to the student. Other variants of KD like Attention Transfer-based KD (ATKD) further use intermediate outputs of teacher's convolution layers while training the student~\cite{atkd}. Of note, KD-based techniques have also been proposed for RNNs~\cite{seqKD,rnnKD}. However, our focus in this paper is on KD-based distributed inference for CNNs.

\subsection{Network Science Background}
Network science has emerged for problems such as modeling information diffusion in social networks, epidemics, transportation, \textit{etc.}~\cite{networksci}. However, to our knowledge, we are the first to utilize network science concepts for model compression. Below, we describe the core ideas from network theory used in this paper.

\subsubsection{\textbf{Node Degree/Hubs}}
Given an undirected network $\mathcal{G}$ with nodes $\mathcal{V}$, links $\mathcal{E}$, and the adjacency matrix $\mathcal{F}=\{F_{ij}\}$ describing link weights between any two nodes $i, j \in \mathcal{V}$. Then, degree $k_i$ of a node $i$ refers to total number of links connected to node $i$. A network can have a \textit{scale-free} structure where some nodes have many connections but most nodes have low degree~\cite{networksci}; such nodes with many connections in scale-free networks act as \textit{hubs} of information. To illustrate, Fig.~\ref{kdb}b shows nodes A, B, and C as examples of hubs in a network. As evident, these nodes are characterized by more connections than other nodes in the network.

\subsubsection{\textbf{Community Structure}}
Many real world networks are organized into groups of tightly connected nodes known as the \textit{community structure}~\cite{newmanComm}. For instance, in a social network, a community can refer to a group of users with common interests like sports or politics. Formally, a community can be defined as a group of nodes for which the number of connections within the group is significantly higher than what one would expect at random. Fig.~\ref{kdb}b shows the community structure in a network. In contrast to older graph partitioning techniques, community detection automatically partitions the network into its natural groupings without requiring us to specify predefined parameters such as number/size of communities. Size and number of communities in a network instead depend on a user-specified \textit{resolution} parameter $\gamma$.

\section{Motivation}
We now discuss the impact of horizontally splitting a Convolutional Neural Network (CNN) on distributed inference performance. More precisely, when a CNN does not fit on a single memory-constrained device, it must be split ``horizontally'' for parallel execution on multiple devices (for better utilization of all resources). However, as mentioned in Section 1, this can result in significant communication costs when each split is deployed on a separate device. To verify this assumption, we conduct the following experiment.

\begin{figure}[]
\centering
\includegraphics[width=0.55\textwidth]{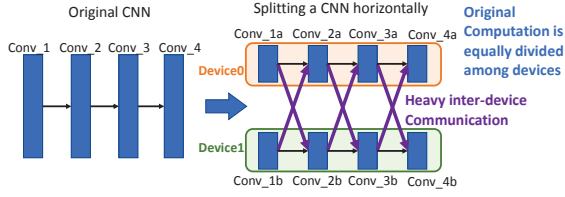}
	\caption{Splitting a deep network horizontally leads to huge communication cost at every step since the next layer convolutions require access to all the channels at the last layer. Original computation is equally divided between the two devices.}
\label{mot}
\end{figure}

We start with a large Wide Resnet (WRN40-4) teacher model with 8.9 million parameters, trained on CIFAR-10 dataset\footnote{The base WRN architecture consists of three groups: $G_0$, $G_1$, and $G_2$ with width (\textit{i.e.}, $\#$channels) in each group as $[16,32,64]$ respectively. Width multiplier is used to increase the number of channels per group~\cite{wrn}; WRN40-4 implies 40 layers and a width multiplier of 4 which results in $[64,128, 256]$ channels per group (see Fig.~\ref{nonn31}a).}~\cite{wrn}. This model takes about 86ms for inference on a single powerful x86 machine. Next, we split the WRN40-4 model horizontally (\textit{e.g.}, if a layer has 32 channels, we split it into two parts with 16 channels each). We then deploy the individual horizontal splits on two powerful x86 machines connected via a wired point-to-point connection. As shown in Fig.~\ref{mot}, since each successive convolution layer requires input from all the output channels from the previous layer, we need to make the output from all devices available to all other devices. Clearly, this results in a significant communication overhead. This is why, even though the computation happens on two powerful x86 machines, the inference time increases \textbf{from 86ms to 1006ms} ($\bm{> 10\times}$ increase in latency). Hence, this experiment shows that, when a deep network does not fit on an IoT-device, splitting it horizontally is simply not an option. Obviously, the problem gets exacerbated for edge devices which often operate on low frequencies and, hence, take even longer for computation.

As a result, a new memory- and communication-aware model compression technique is imperative for distributed inference on IoT-devices. Therefore, we next describe our proposed approach.

\section{Proposed Approach}
We first formulate our problem (Section 4.1), followed by our proposed solution which consists of two stages: (1) Network science-based knowledge partitioning of the teacher: We find disjoint partitions of teacher's knowledge which lead to independent students with minimal communication costs (Section 4.2, Fig.~\ref{flow}b). (2) Proposed NoNN architecture and the training process: We first select an efficient model architecture for each student that reduces its memory and FLOPS, and then jointly train individual NoNN students on disjoint partitions from teacher (Section 4.3; Fig.~\ref{flow}c).

\subsection{Problem Formulation}
In this section, we formulate teacher's knowledge partitioning subject to the resource-constraints of individual students.

\begin{figure*}[t]
\centering
\includegraphics[width=\textwidth]{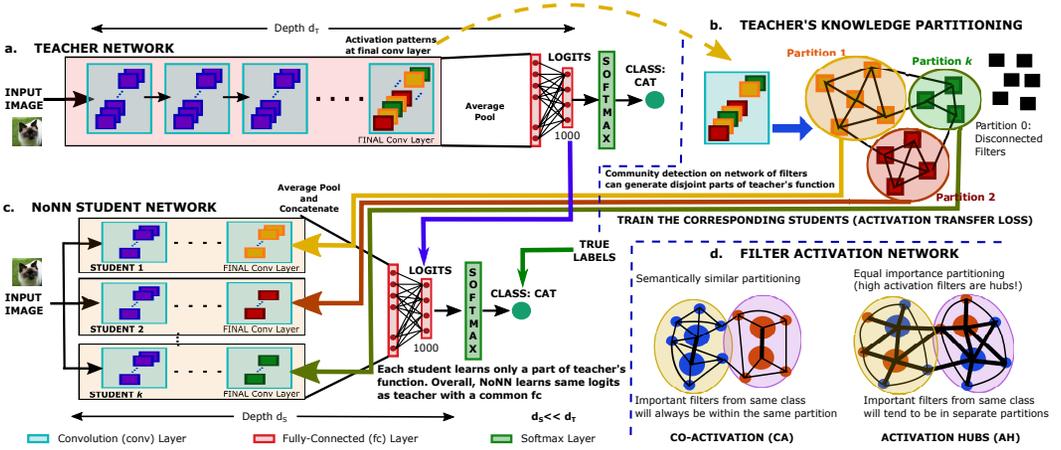}
	\caption{Complete flow of our approach. (a) The pretrained teacher model exhibits certain activation patterns for the validation set. (b) We use these activation patterns to build a network of filter activity. Community detection on the filter network partitions teacher's knowledge into simpler functions. (c) Train individual students in NoNN to mimic each partition. (d) Teacher's knowledge can be partitioned in different ways depending on how the filter activation network is built (\textit{e.g.}, via Co-Activation (CA) rule or Activation Hubs (AH) rule). Node colors represent filters relevant to a given class, whereas node sizes represent importance of a filter (determined by activation value).}
\label{flow}
\end{figure*}

As shown in Fig.~\ref{flow}a, the output of final convolution (fconv) layer of teacher ($T_{fconv}$) gets average-pooled and passes through the fully-connected (fc) layer to yield \textit{logits}. The logits then pass through the softmax layer to generate probabilities. Since our goal is to break teacher's learned function into multiple disjoint parts, we focus on teacher network's fconv. Specifically, we partition the teacher's fconv layer by looking at the \textit{patterns of activation of filters} as validation images pass through the teacher (see Section 4.2).

Partitioning a teacher's fconv often uncovers a partition $P_0$ which does not contribute to teacher's validation accuracy. Therefore, we should find and remove the largest set of filters $P_0 \subset T_{fconv}$ that does not contribute to teacher's accuracy since such filters only transfer noise to the students during distillation (Section 4.2). Moreover, the sizes of partitions of $T_{fconv}$ should be almost equal so that the resulting student networks fit within a memory budget $\mathcal{B}_{mem}$, while their computation costs fit a FLOP budget $\mathcal{B}_{FLOP}$. Let $h: \mathbb{R}\rightarrow \mathbb{R}$ compute the memory of student $i$ depending on its partition size $|P_i|$, and $f: \mathbb{R}\rightarrow \mathbb{R}$ compute the resulting FLOPS. Then, the \textit{knowledge-partitioning problem} is expressed as:
\begin{equation}
	\begin{aligned}
        & \underset{\mathcal{P}=\{P_0, P_1, \ldots, P_k\}}{\text{min}}\ \ \ |\Delta \text{val}| - |P_0| &\\
        &\ \ \ \ \text{subject to} \ \ \ \ \ \  \ |P_1| \approx |P_2| \ldots \approx |P_k|,\  P_i \sim \text{Rule } R,\ P_i \subset T_{fconv}\\
        & \ \ \ \ \ \ \ \ \ \ \ \ \ \ \ \ \ \ \ \ \ \ \ \ \ \ \  \ \ \ \ \ \ \ \ \ \ \ \ \  \ \ \ \ \ \ \ \ \ \ \ \ \ \ \ \ \ \ \ \ \ \ \ \ \  \ \ \ \ \ \ \ \ \ \ \ \ \ \  \ 			\forall i \in \{0,\ldots,k\} \\
        & \ \ \ \ \ \ \ \ \ \ \ \ \ \ \ \ \ \ \ \ \ \ \ \ \ \ \ 				  |\Delta \text{val}| < \epsilon &\\
     	&  \ \ \ \ \ \ \ \ \ \ \ \ \ \ \ \ \ \ \ \ \ \ \ \ \ \ \ \text{max}(h(|P_1|), h(|P_2|), \ldots, h(|P_k|)) < \mathcal{B}_{mem} & \\
        & \ \ \ \ \ \ \ \ \ \ \ \ \ \ \ \ \ \ \ \ \ \ \ \ \ \ \  \text{max}(f(|P_1|), f(|P_2|), \ldots, f(|P_k|)) < \mathcal{B}_{FLOPS}  &
    \end{aligned}
\label{fitlerNet}
\end{equation}
where, $|\Delta \text{val}|$ denotes the absolute value of change in validation accuracy due to removal of $P_0$ filters. Also, $|P_i|$ does not denote the absolute value, but rather the number of elements in the set $P_i$ (\textit{i.e.}, the cardinality of $P_i$). The first constraint in~\eqref{fitlerNet} aims to keep the sizes of all partitions (except $P_0$) almost equal since the sizes of student models must be below the \textit{fixed} memory budget $\mathcal{B}_{mem}$. The objective function and the second constraint aim to minimize change in teacher's validation accuracy while removing as many useless filters from $T_{fconv}$ as possible. The last two constraints in problem~\eqref{fitlerNet} specify the per-device memory- and FLOP-budgets that must be satisfied by each student (which will be deployed on individual devices). Note that, while it is always possible to satisfy memory- and FLOP-constraints, there will be a tradeoff between model-size or computation and the NoNN accuracy. For instance, for fixed number of students, larger individual student modules can lead to higher overall accuracy. Finally, the rule $R$ needed to determine how the partitions are formed is described below.

\subsection{Network Science-Based Knowledge Partitioning: Filter Activation Network}
We propose a network science-based solution to the knowledge partitioning problem. The idea is to uniformly distribute knowledge from teacher's fconv using network science in order to jointly train individual students in the NoNN.

To partition teacher's fconv, we first build a network $\mathcal{F}=\{F_{ij}\}$ of filter activation patterns, where $i$ and $j$ are two filters. Now, let $a_i$ denote the \textit{average activity} of a filter $i$ for a given image in the validation set ($\text{val}$). Average activity of a filter $i$ is defined as the averaged output of the corresponding output channel of teacher's fconv. For instance, suppose a teacher network's fconv has 256 output channels, and height and width of its activation map is $8\times 8$ pixels. Then, to obtain average activity of filter $i$, we average the value of its $8\times 8$ output channel activation map. Essentially, we use this average activity metric as a measure of \textit{importance} of a filter for a given class of images; that is, the higher the average activity of a filter, the more important it is for classification for some class of images. In Fig.~\ref{flow}d, each circle depicts a filter at teacher's fconv, and filters shown in same color activate for similar classes. Also, the bigger the size of the circle, the more important is the filter for the corresponding class (\textit{i.e.}, it has higher average activity).

To create the filter activation network, there exist multiple strategies (\textit{i.e.}, rules $R$) by which the filters can be connected. Fig.~\ref{flow}d illustrates two such strategies: \textbf{(\textit{i}) Activation Hubs (AH):} Encourage connections between very important and less important filters (\textit{i.e.}, less number of high-importance filters get surrounded by many low-importance filters and thus act as hubs), and \textbf{(\textit{ii}) Co-Activation (CA):} Filters that activate together for similar classes get connected. The former rule encourages partitions of roughly equal importance (since each partition will have a few important filters). The latter rule partitions the filters into semantically similar parts. Formally,
\begin{equation}
F_{ij} = \begin{cases}
\sum_{\text{val}}a_i a_j|a_i - a_j| &\textbf{AH }\text{Rule}\\
\sum_{\text{val}}\frac{a_i a_j}{|a_i - a_j|+1} &\textbf{CA }\text{Rule}
\end{cases}
\end{equation}

Let us analyze what happens during the AH case. If $i$ and $j$ are two filters with average activities $a_i$ and $a_j$, respectively, the $F_{ij}$ link weight for AH case will be high if, for a given image, either $a_i$ is high and $a_j$ is low, or vice versa. Note that, if either $a_i$ or $a_j$ is close to zero, that link is not created. Further, when $a_i$ and $a_j$ are both very high for the same image (\textit{i.e.}, both filters are important for same class), then that link will also not be encouraged (since the link weight will be low due to the $|a_i-a_j|$ part). Therefore, important filters from the same class will be discouraged from connecting together. Consequently, for AH Rule, highly important filters (with high $a_i$) will be encouraged to connect with not-so-important filters (from same or other classes). Important filters from same classes will be forced to occupy separate communities; hence, the knowledge is distributed uniformly across the students. Similar discussion holds for CA Rule where the communities represent semantically-similar features. Of note, this flexibility in partitioning teacher's fconv comes from network science ideas.

Of course, there can be other ways filters may be connected. Indeed, if the dataset contains too many semantically similar classes, CA can lead to heterogeneous student models (because many filters from $T_{fconv}$ will go into a few partitions). Hence, to obtain student models of similar sizes, we only explore AH in this paper. In matrix form, adjacency matrix of AH network can be written as follows:
\begin{equation}
\mathcal{F}^{AH} = \sum_{n \in \text{val}} a_n a_n^T \odot |\mathbf{D}_n- \mathbf{D}_n^T|, 
\end{equation}
where, $a_n$ is the vector of average filter activities ($a_i$'s) for each image $n \in \text{val}$, matrix $\mathbf{D}_n$ contains all columns as $a_n$, and $\odot$ denotes element-wise multiplication. To partition this AH network, we need to detect communities by maximizing a modularity function as explained in the network science literature~\cite{newmanComm}:
\begin{equation}
    \max_{\mathbf{g}=\{g_0, g_1, \ldots, g_{l-1}\}}\ \  \frac{1}{2m} \sum_{ij} \Bigg[F_{ij}^{AH} - \frac{1}{\gamma} \cdot \frac{k_i k_j}{2m}\Bigg]\delta(g_i, g_j), 
    \label{comm}
\end{equation}
where, $m$ is $\#$edges, $k_i$ is degree (number of connections) of node $i$, resolution $\gamma$ controls the size/number of communities, and $\delta$ is Kronecker delta. The idea is to find groups of tightly connected nodes, $\mathbf{g}=\{g_0, g_1, \ldots, g_{l-1}\}$, which map the nodes $\mathcal{V}$ to $l$ communities. Finally, these $l$ communities are converted to $k$ partitions $P_i$ by combining the communities such that the partition sizes are almost equal (so that resulting students stay within the budget). This partitioning $\mathcal{P}$ is then used to train individual student networks. Of note, removing community $g_0$ (which consists of all disconnected nodes in the filter network) often has very little impact on teacher's accuracy. Hence, partition $P_0 = g_0$ is obtained directly by community detection. Consequently, we solve problem~\eqref{fitlerNet} by detecting and removing the largest community $g_0$ that does not change teacher's validation accuracy, and then combine the rest of the communities to form the remaining partitions. For instance, say we have four remaining communities after removing $g_0$: $\{|g_1|=12,|g_2|=14,|g_3|=25, |g_4|=30\}$ ($|x|$ indicates size of group $x$), and we want to combine them into two partitions of nearly equal sizes. Then, we can combine them as $\mathcal{P}=\{|P_1|=|g_2\cup g_3|=39, |P_2|=|g_1\cup g_4|=42\}$. Hence, we obtain the required $k$ almost equal partitions from $l$ communities.

\subsection{Network of Neural Networks (NoNN)}
Once we have partitioned the teacher's fconv, we jointly train our student networks on individual partitions from $T_{fconv}$ as shown in Fig.~\ref{flow}c. However, we must first make the following design decisions: (\textit{i}) Select the deep network architecture for individual students, and (\textit{ii}) Select how students are connected together in NoNN. 

\subsubsection{\textbf{Student Architecture Selection}}
Similar to prior art in KD~\cite{hintonKD, atkd}, we select our individual student networks to be significantly reduced versions of the teacher (\textit{i.e.}, far fewer layers and lower width). Specifically, we pick our individual student models based on user-defined memory- and FLOP-constraints for IoT-devices. For instance, for CIFAR-10 dataset, our teacher is Wide Resnet model WRN40-4 with 40 layers and width multiplier of 4. That is, as shown in Fig.~\ref{nonn31}a (left), it has 64 channels in first group, 128 in second group, and 256 in third group. Also, suppose that our each individual student must have less than 500K parameters. Then, we select our individual student (see Fig.~\ref{nonn31}a (right)) with 16 layers and channel-width appropriately adjusted so that each student has less than 500K parameters. Note that, student networks are not exactly the same as they mimic different partitions. Therefore, we need an additional $1\times 1$ convolution layer to make the dimensions equal between the teacher's knowledge partition and individual student's output layer. Next, we describe the architecture of NoNN containing multiple such students, and the training process.

\subsubsection{\textbf{NoNN Architecture}}
\begin{figure}[tb]
\centering
\includegraphics[width=0.6\textwidth]{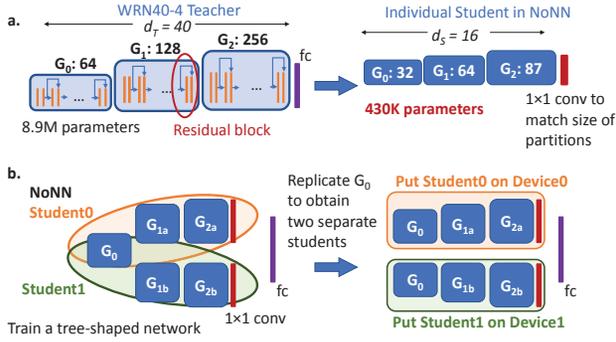}
	\caption{(a) Selecting an individual NoNN student architecture: Reduce depth and channel-width so that it fits within the desired budgets. (b) Overall, NoNN trains a tree-shaped network (where possible) with some initial layers common (denoted as $\bm{G_0}$). At inference time, replicate the common group $\bm{G_0}$ and put individual students on separate devices.}
\label{nonn31}
\end{figure}

A NoNN consisting of two students is shown in Fig.~\ref{nonn31}b. As evident, wherever possible, we make the initial few layers from the students common (since the knowledge learned at initial few layers will be common for all students). 
Moreover, at inference time, the common layers between all students can be simply replicated across multiple devices without significant increase in memory per student (since the common group is not too big). Once the common group is replicated, the individual students are completely independent and do not communicate until the final fc layer (see Fig.~\ref{nonn31}b). Hence, we obtain a final NoNN architecture which consists of multiple disjoint students. The output from all students is concatenated and passed through a fc to yield logits. 

\subsubsection{\textbf{Training Loss}}
To train the overall NoNN, we use the KD loss (see Eq.~\eqref{kdloss1}) and, as given below, we additionally propose a new loss function called \textit{activation-transfer} for transferring partitioned knowledge from the teacher to individual students. The main motivation behind the activation-transfer loss function is that each individual student must mimic the corresponding partition of teacher's knowledge. Specifically, for each partition of teacher's knowledge, we want to minimize the error between activations of the teacher's filters (that belong to the given partition) and activations of filters in the corresponding student. We define the activation-transfer loss function as given below:
\begin{equation}
\mathcal{L}^{Act}(\theta_S)=\sum_{p \in \mathcal{P}}\Big\| \frac{v^F_T(p)}{||v^F_T(p)||}- \frac{v^F_S(p)}{||v^F_S(p)||} \Big\|_2^2
\label{lAct}
\end{equation}
where, $v^F_Z(p)=vec(A^F_Z(p))$, $Z \in \{T, S\}$ is the vectorized fconv activations for partition $p \in \mathcal{P}$ of teacher or for individual student network in NoNN, and the $||v^F_Z(p)||$, $Z \in \{T, S\}$ terms denote the partition-wise normalization. Hence, the total loss is given by $\mathcal{L}=\mathcal{L}^{kd}(\theta_S)+\beta \mathcal{L}^{Act}(\theta_S)$ which can be minimized via stochastic gradient descent.

This completes our proposed NoNN. We next present detailed experimental evaluation for NoNN.

\section{Experimental Setup and Results}
In this section, we first present our experimental setup and then results for different datasets and NoNN architectures.

\subsection{Experimental Setup}
We compress various deep networks for five well-known image classification tasks: CIFAR-10, CIFAR-100, Scene~\cite{scen}, Caltech-UCSD-Birds (CUB)~\cite{cub}, and Imagenet. Scene and CUB datasets belong to the transfer learning domain where a pretrained model is finetuned on a different problem. For comparison, we consider strong teacher-student baselines like KD~\cite{hintonKD}, Attention-transfer with KD (ATKD) \cite{atkd}. For KD and ATKD, we use models from the WRN family~\cite{wrn} as single, large students. Also, \textit{total parameters} in NoNN (\textit{i.e.}, parameters in all students combined) are comparable to the parameters in KD/ATKD baseline models. We also show that we outperform prior models like Splitnet~\cite{splitnet}. Further, we show experiments with up to 8 student networks for CIFAR-10/100. With these experiments, we thoroughly demonstrate that NoNN compresses the teacher model into disjoint subsets which do not communicate until the fc layer.

We first train the teacher model on $90\%$ of training dataset while saving $10\%$ for validation. This validation set is used for building the filter activation network to partition the teacher's knowledge. Then, we use the Activation Transfer loss given in Eq.~\eqref{lAct} and KD-loss to train the compressed NoNN model. Moreover, all the accuracies reported in the paper are on test set for the respective datasets. We set the number of partitions, $k=2$. To build more than two students, since our objective is to keep the individual student architectures as similar as possible, we simply shuffle the filters in these two partitions while transferring knowledge to rest of the students. This ensures that the size of students is not too diverse and limits the possible factors that can contribute to the improvement in accuracy of our model. Also, we set $\alpha=0.9$ and selected the best $\beta=\{10^1,10^2,\ldots,10^4\}$ for both ATKD~\cite{atkd} and NoNN. For training NoNN, we set the initial learning rate as $0.1$ and used a momentum of $0.9$. Finally, our entire framework is implemented in Pytorch which is run on NVIDIA GTX 1080-Ti and Titan Xp GPUs.

\subsection{Results} \label{sec:results_NoNN}
We first show results on CIFAR-10/100 and transfer learning datasets (Scene, CUB), and then our preliminary results for Imagenet.

\subsubsection{\textbf{CIFAR-10}}
\begin{figure*}[!t]
\centering
\includegraphics[width=0.9\textwidth]{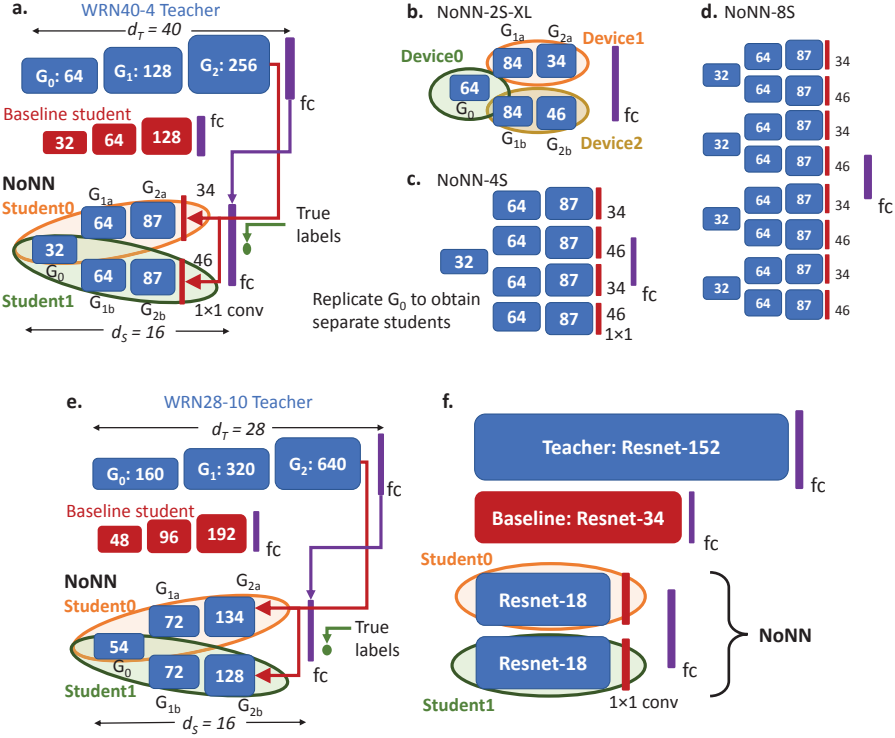}
	\caption{Teacher, baseline student, and NoNN models for various datasets. (a) WRN-40-4 teacher, WRN-16-2 baseline, and 2-student NoNN for CIFAR-10. (b) A larger 2-student NoNN (NoNN-2S-XL) can be distributed on three separate devices to keep FLOPS under a budget. (c, d) 4-student and 8-student NoNNs. (e) CIFAR-100 models, (f) Transfer learning models. After replicating the common groups (if any), the individual students in our proposed NoNN will communicate only at fc layer.}
\label{arch}
\end{figure*}

Fig.~\ref{arch}a depicts our teacher model for CIFAR-10 as WRN40-4 and our baseline student for KD and ATKD as WRN16-2. As evident, knowledge from partitions of only 34 and 46 filters from teacher's fconv is transferred to student-0 and student-1, respectively (via $1\times 1$ convolutions). The rest 176 out of 256 filters at teacher's fconv do not contribute to teacher's accuracy and, hence, form partition $P_0$ (or community $g_0$) in $\mathcal{F}^{AH}$; this partition is not used for transferring knowledge to NoNN students. We set $\mathcal{B}_{FLOPS}\approx 200$M which is close to baseline FLOPS used by WRN16-2 model. Further, based on our initial motivation w.r.t. Arm Cortex-M, $\mathcal{B}_{mem}=0.5$M parameters (since for models quantized to 8-bits, $0.5$M parameters mean less than $500$KB of memory).
\begin{table}[!t]
\centering
\caption{CIFAR-10 Teacher-Student Results$^*$}
\scalebox{0.8}{
\label{c102S}
\begin{tabular}{|l|c|c|c|} 
\hline
Model& $\#$parameters & $\#$FLOPS & Accuracy\\
& (largest student) & (largest student)& \\
\hline \hline
Teacher WRN40-4& $8.9$M& $2.6$G& $95.49\%$\\ \hline
KD WRN16-2~\cite{hintonKD}& $0.7$M& $202$M& $93.79\pm 0.15\%$\\ \hline
ATKD WRN16-2~\cite{atkd}& $0.7$M& $202$M& $93.83\pm 0.08\%$\\ \hline
\textbf{NoNN-2S}& $\bm{0.43}$\textbf{M}$^{**}$& $\bm{167}$\textbf{M}& $\bm{94.32\pm 0.17\%}$\\ \hline \hline
KD: NoNN-2S-XL& $0.46$M& $245$M& $93.96\pm 0.09\%$\\ \hline
\textbf{NoNN-2S-XL}& $\bm{0.46}$\textbf{M}$^{**}$& $245$M$^{***}$& $\bm{94.53\pm 0.19\%}$\\ \hline
\end{tabular}
}
\begin{flushleft}
{\scriptsize $^{*}$For statistical significance, results are reported as mean $\pm$ standard deviation of five experiments.}\\
{\scriptsize $^{**}$All NoNN $\#$parameters/FLOPS are reported for one student only. Complete NoNN has very similar number of parameters as baselines for fair comparison. Our contribution is to show that NoNN can be broken down into disjoint parts that stay below certain budgets and do not communicate until the last layer.}\\
{\scriptsize $^{***}$NoNN-2S-XL can be distributed onto three devices as shown in Fig.~\ref{arch}b.}
\end{flushleft}
\end{table}

As shown in Table~\ref{c102S}, our NoNN-2S model (Fig.~\ref{arch}a) achieves higher accuracy than the baseline student model. Note that, each student in NoNN utilizes fewer FLOPS (\textit{i.e.}, $167M$ FLOPS) than the allowed budget. Moreover, although \textit{each} student has $0.43$M parameters, due to common $G_0$, total parameters for NoNN-2S in Fig.~\ref{arch}a is $0.82$M (comparable to $0.7$M in WRN16-2). Further, \textit{after replicating $G_0$}, student-0 and student-1 will not communicate with each other until the final fc layer. Clearly, when quantized, our individual student models of $430$K parameters can fit on a device with $500$KB memory ($430$K parameters are after replicating $G_0$). Indeed, if a device cannot fit more than $500$K parameters, prior models such as WRN16-2 (with $700$K parameters) will need to be distributed across multiple devices which will lead to communication at every layer; that is, when distributed, the baseline WRN16-2 will communicate at all 16 layers, whereas NoNN communicates only at the final layer. 

This situation gets exacerbated by higher accuracy baseline student models (\textit{e.g.}, 40-layer WRN40-2) for which communication costs at every layer grow rapidly. In contrast, as we shall see shortly, our student models can be very flexible: we can have larger students (see Fig.~\ref{arch}b), or many disjoint students without significantly increasing the communication costs (\textit{e.g.}, see NoNN-4S and NoNN-8S in Fig.~\ref{arch}(c, d)). Concrete energy and latency for CIFAR-10 experiments are reported in the hardware deployment described in Section 6.

For an even higher per-device FLOP budget (say, $250$M), a NoNN-2S-XL student architecture can be distributed on three devices (see Fig.~\ref{arch}b). Although the larger student in this case has $0.46$M parameters ($G_0\rightarrow G_{1b} \rightarrow G_{2b}$), due to a bigger common $G_0$ group, total parameters for NoNN-2S-XL is $0.77$M (again, comparable to $0.7$M in WRN16-2). Also, each individual student in this model has higher FLOPS because the common group $G_0$ is wider and has $245$M FLOPS. Since it is impractical to replicate this group, as shown in Fig.~\ref{arch}b, we can assign a completely new device (device-0) to this group and put $G_{1a}\rightarrow G_{2a}$ ($G_{1b}\rightarrow G_{2b}$) on device-1 (device-2). 

Table~\ref{c102S} demonstrates that our NoNN-2S-XL achieves even better accuracy (<$1\%$ away from the teacher). Moreover, this increase in accuracy is not only due to an increased FLOP budget. To demonstrate, we show that filter network community-based knowledge transfer from teacher plays a more significant role in accuracy improvement. If we simply train our NoNN-2S-XL student model via KD, \textit{i.e.}, without the $\mathcal{L}^{Act}$ loss defined in Eq.~\eqref{lAct}, the accuracy increases only slightly over the baseline WRN16-2 (\textit{e.g.}, $93.96\%$ \textit{vs.} $93.79\%$). Therefore, community-based knowledge partitioning is extremely important for successfully training our proposed NoNN. 

Next, we show that NoNN is scalable to many datasets and model sizes for different memory and FLOP budgets. 

\subsubsection{\textbf{CIFAR-100 and Comparison with SplitNet}}
Table~\ref{c1002S} shows results for CIFAR-100. Here, the teacher is a WRN28-10 model with $36.5$M parameters and our baseline student is WRN16-3 model containing $1.5$M parameters (see Fig.~\ref{arch}e). Therefore, we create a NoNN with overall $1.5$M parameters as shown in Fig.~\ref{arch}e; this NoNN can be split into two students where the larger student has $0.85$M parameters (after replicating $G_0$). Also, the FLOPS for our larger student are lower than the WRN16-3 baseline. Again, NoNN-2S model is found to be more accurate than KD and ATKD, while creating disjoint students that can fit within some memory budget (say, $1$MB). Once our disjoint students fit within the $1$MB budget, techniques like MoDNN~\cite{modnn} can further distribute our model's computation on more devices with similar memory-constraints to reduce the FLOPS. MoDNN assumes that the model fits on the device, and our work makes it possible to meet the memory-constraints.
\begin{table}[tb]
\centering
\caption{CIFAR-100 Teacher-Student Results$^*$}
\scalebox{0.8}{
\label{c1002S}
\begin{tabular}{|l|c|c|c|} 
\hline
Model& $\#$parameters & $\#$FLOPS & Accuracy\\
& (largest student) & (largest student)& \\
\hline \hline
Teacher WRN28-10& $36.5$M& $10.48$G& $79.28\%$\\ \hline
KD WRN16-3~\cite{hintonKD}& $1.5$M& $446$M& $73.99\pm0.26\%$\\ \hline
ATKD WRN16-3~\cite{atkd}& $1.5$M& $446$M& $74.90\pm0.17\%$\\ \hline
\textbf{NoNN-2S}& $\bm{0.85}$\textbf{M}& $\bm{345}$\textbf{M}& $\bm{75.74\pm 0.31\%}$\\ \hline
\end{tabular}
}
\begin{flushleft}
{\scriptsize $^{*}$For statistical significance, results are reported as mean $\pm$ standard deviation of five experiments.}
\end{flushleft}
\end{table}

We next compare our model to SplitNet~\cite{splitnet} that splits a deep network into multiple disjoint parts (albeit without any model compression ideas such as resource-constraints). Kim \textit{et al.} obtained two CIFAR-100 models via SplitNet: (\textit{i}) A $7.42$M parameter WRN model with 2 sub-groups which achieves $76.04\%$ accuracy, and (\textit{ii}) A $4.12$M parameter model with 4 sub-groups which achieves $75.2\%$ accuracy. For fair comparison with SplitNet, we create two corresponding models: First with 2-students ($7.42$M parameters), and the second with 4-students ($4.13$M parameters), respectively. While keeping the experimental setup same as SplitNet (\textit{e.g.}, same validation set size, \textit{etc.}), we achieve $\bm{79.07\%}$ \textbf{accuracy} for the $7.42$M parameter model, and $\bm{77.42\%}$ \textbf{accuracy} for the $4.13$M parameter model. Therefore, the proposed NoNN significantly outperforms SplitNet for similar model-sizes. Moreover, the NoNN-2S model shown in Table~\ref{c1002S} which has merely $1.5$M total parameters (\textit{i.e.}, $0.85$M parameters per student) achieves higher accuracy than the $4.13$M parameter SplitNet model.

\subsubsection{\textbf{Varying Number of Students}}
Now, we vary students from two to eight for CIFAR datasets. As evident from Table~\ref{numStud}, NoNN models achieve close to teacher's accuracy while reducing memory/FLOPS by orders of magnitude (\textit{i.e.}, $2.5\times$-$24\times$ reduction in $\#$parameters and $2\times$-$15\times$ reduction in FLOPS w.r.t. teacher). Specifically, the NoNN-8S model in Fig.~\ref{arch}d achieves $95.02\%$ accuracy (\textit{i.e.}, less than $0.5\%$ away from teacher) while using eight separate students (after $G_0$ is replicated), each of which can fit within $167$M FLOPS and $430$K parameters (\textit{i.e.}, <500KB when quantized to 8-bits), and do not communicate until the final fc layer. Overall, this results in $2600/(167\times 8)\approx 2\times$ reduction in FLOPS, and $8.9/(0.43\times 8)\approx 2.5\times$ reduction in parameters over the teacher. 

Finally, to compare NoNN-8S for CIFAR-10, we also experimented with a larger WRN40-2 student (40 layers, $2.2$M parameters, and $655$M FLOPS) with ATKD~\cite{atkd}. Although, this model achieves similar accuracy of $95.03\%$, it clearly does \textit{not} result in an architecture that can be distributed on multiple devices (since it will require $2.2$MB storage when quantized to 8-bits). As a result, distributing this single model on multiple devices will lead to heavy inter-device communication at each of the 40 layers (see Section 3),  whereas our model will communicate only at the final layer. This clearly highlights the significance of our proposed NoNN.
\begin{table*}
\centering
\caption{Accuracy for more students$^*$}
\scalebox{0.76}{
\label{numStud}
\begin{tabular}{|c||c|c|c||c|c|c||c|c|c|} \hline
$\#$S &\multicolumn{6}{c||}{CIFAR-10}& \multicolumn{3}{c|}{CIFAR-100}\\
\cline{2-7} \cline{8-10}
 &\textbf{NoNN}&$\Delta$M$^{**}$&$\Delta$F$^{**}$&\textbf{NoNN-XL}&$\Delta$M&$\Delta$F&\textbf{NoNN}&$\Delta$M&$\Delta$F\\ \hline \hline
$2$& $94.32\pm 0.17\%$& $10\times $& $7.7\times$ & $94.53\pm 0.19\%$& $11\times$ & $5\times$& $75.74\pm 0.31\%$ & $24\times$ & $15\times$\\ \hline
$4$& $94.64\pm 0.16\%$& $5.1\times$ & $3.9\times$ & $94.87\pm 0.14\%$& $6.3\times$ & $3.3\times$ & $77.07\pm 0.18\%$ & $12\times$ & $7.6\times$\\ \hline
$6$& $94.90\pm 0.06\%$& $3.5\times$ & $2.6\times$ & $95.02\pm 0.10\%$& $3.9\times$ & $1.7\times$ & $\bm{77.25\pm 0.24\%}$ & $7.7\times$ & $5\times$\\ \hline
$8$& $\bm{95.02\pm0.08\%}$& $2.5\times$ & $2\times$ & $\bm{95.28\pm 0.07\%}$& $2.9\times$ & $1.3\times$ & $77.03\pm 0.17\%$ & $5.7\times$& $3.8\times$\\ \hline \hline
T& $95.49\%$& $1\times$ & $1\times$ & $95.49\%$& $1\times$ & $1\times$ &  $79.28\%$ & $1\times$ & $1\times$  \\ \hline
\end{tabular}
}
\begin{flushleft}
{\scriptsize $^{*}$For statistical significance, results are reported as mean $\pm$ standard deviation of five experiments.}\\
{\scriptsize $^{**}\Delta$M ($\Delta$F) shows compression rate, \textit{i.e.}, $\Delta$Memory ($\Delta$FLOPS) for \textit{complete} NoNN (\textit{i.e.}, parameters of all students combined) w.r.t. teacher.}
\end{flushleft}
\end{table*}

\subsubsection{\textbf{Transfer Learning Datasets}}
We now demonstrate the effectiveness of NoNN for transfer learning tasks where the idea is to finetune a model pretrained on Imagenet w.r.t. a new dataset. Table~\ref{tl2S} shows results for two such datasets: (\textit{i}) Indoor Scene Classification, (\textit{ii}) Caltech-UCSD Birds. For both datasets, we used Resnet-152 as teacher model, Resnet-34 as baseline student, and our NoNN consists of two separate Resnet-18 models with a common fc layer (see Fig.~\ref{arch}f). As evident from Table~\ref{tl2S} and Fig.~\ref{arch}f, our NoNN-2S model (total $23$M parameters) consists of disjoint parts of $11$M parameters each, which do not communicate until the last layer. Also, for Scene dataset, NoNN-2S achieves slightly lower (but still comparable) accuracy than ATKD on Resnet-34. However, NoNN-2S achieves much higher accuracy than prior approaches for CUB. 
\begin{table*}[tb]
\centering
\caption{Transfer Learning Results$^*$}
\scalebox{0.76}{
\label{tl2S}
\begin{tabular}{|l|c|c||c|c||c|c|} 
\hline
Model& $\#$parameters& $\#$FLOPS & \multicolumn{2}{c||}{Accuracy}&$\Delta$M$^{**}$&$\Delta$F$^{**}$\\
\cline{4-5}
& (largest student) & (largest student)& Scene& CUB & &\\
\hline \hline
Teacher Resnet-152& $58$M& $11$G& $79.44\%$& $80.94\%$ & $1\times$ & $1\times$\\ \hline
KD Resnet-34~\cite{hintonKD}& $22$M& $4$G& $76.92\pm 0.39\%$& $78.59\pm 0.22\%$ & $2.5\times$ & $2.8\times$\\ \hline
ATKD Resnet-34~\cite{atkd}& $22$M& $4$G& $\bm{77.63\pm 0.94\%}$& $79.10\pm 0.56\%$ & $2.5\times$ & $2.8\times$\\ \hline
\textbf{NoNN-2S}& $\bm{11}$\textbf{M}& $\bm{2}$\textbf{G}& $\bm{77.48\pm 0.76\%}$& $\bm{79.81\pm 0.33\%}$ & $2.5\times$ & $2.8\times$ \\ \hline
\end{tabular}
}
\begin{flushleft}
{\scriptsize $^{*}$For statistical significance, results are reported as mean $\pm$ standard deviation of five experiments.}\\ 
{\scriptsize $^{**}\Delta$M ($\Delta$F) shows compression rate, \textit{i.e.}, $\Delta$Memory ($\Delta$FLOPS) for \textit{complete} NoNN (\textit{i.e.}, parameters of all students combined) w.r.t. teacher.}
\end{flushleft}
\end{table*}

\subsubsection{\textbf{Preliminary Imagenet Results}}
We used the fastai setup \cite{fastai} to obtain imagenet results in a fast and inexpensive way. Our teacher model is Resnet-34 with $73\%$ top-1 and $91\%$ top-5 accuracy. We used Resnet-18 model ($11.68$M parameters) as a baseline ATKD student which achieved $69.98\%$ top-1 and $89.50\%$ top-5 accuracy. Our NoNN consists of two students (each with around $6$M parameters; total $11.69$M parameters). For training our NoNN, we use attention transfer losses~\cite{atkd} on \textit{each} student since activation transfer loss led to some underflow problems\footnote{Fastai uses a floating point-16 (fp16) format for computation on 8 NVIDIA Volta GPUs. Due to an underflow problem on fp16 (which is well-known to happen with fp16 format), we couldn't use activation transfer $\mathcal{L}^{Act}$ losses for training our NoNN.}. Even with this setup, our NoNN achieved \textit{comparable} accuracies, \textit{i.e.}, $69.82\%$ top-1, and $89.61\%$ top-5 while ensuring that the two students do not communicate until the last layer. With availability of more resources, we should be able to improve these results even further. 

So far, we have shown that NoNN achieves higher accuracy than several baselines, and minimal communication among students. We have also demonstrated that NoNN achieves close to teacher's accuracy with $2.5\times$-$24\times$ lower memory, and  $2\times$-$15\times$ fewer FLOPS than the teacher model. Next, we deploy our models on real edge devices and show significant gain in performance and energy.

\section{Case Study: Hardware Deployment}
In this section, as a case study, we deploy our NoNN models on two hardware-constrained devices to quantify their effectiveness in terms of latency and energy on real hardware.

\subsection{Hardware Setup}
We implement NoNN on edge devices such as Raspberry Pi (RPi) and Odroid-XU4S. We demonstrate an extensive exploration of two scenarios that can benefit from our approach: (\textit{i}) Homogeneous case in which all devices are identical, and (\textit{ii}) Heterogeneous case, where devices have different resources and computing power. We perform detailed evaluation of accuracy, performance, and energy for each scenario. We show these results for our teacher (WRN40-4), NoNN-2S, and NoNN-8S models trained on the CIFAR-10 dataset.

We use eight Raspberry Pi-3 Model-B and eight Odroid-XU4S boards. Each RPi has an Arm Cortex-A53 quad-core processor with nominal frequency of 1.2GHz and 1GB of RAM. We scaled the frequency of RPi's down to 400MHz as the RPi's running at maximum frequency demonstrated unstable behavior and would often crash due to high temperature. The XU4S boards have a Samsung Exynos5422 SoC, which has an Arm big.LITTLE architecture with 4 big Cortex-A15 and 4 little Cortex-A7 cores, and 2GB of RAM. The Odroid is executed at the nominal frequency: 2GHz for the big cores and 1.4GHz for the little cores. To maximize performance, we run the inference on the big cores in the Odroids. 

In order to measure voltage, current, and power, we use the Odroid Smart Power 2 for the Odroid-XU4S and AVHzY USB Power Meter for the RPi's. In addition, we use an x86 machine (Core i7), further referred as \textit{server}, to send the images to the devices, concatenate the results, and apply the fc layer. Of note, the fc is small and could also be placed in another edge/mobile device. 
We perform all of our experiments with a point-to-point wired local network connecting all devices together.

Since the NoNN models were built with PyTorch, we use TVM\footnote{TVM Compiler: \url{https://tvm.ai/}}, a deep learning compiler that optimizes NN models for several hardware architectures, to deploy the trained model into the RPi's and Odroid boards. To achieve this, the PyTorch models are converted into an intermediate representation with Open Neural Network Exchange (ONNX) \cite{ONNX}. Then the ONNX representation is used by TVM to generate the binary that is deployed in the target device. In addition, we integrate these binaries into a TCP-wrapper so that each student can be distributed to separate devices and the communication is performed through the TCP protocol.

\subsection{Hardware Results}
We first show the results for NoNN when all devices are identical (\textit{i.e.}, the homogeneous case). Then, we address the heterogeneous scenario, where one device is more powerful than the rest. We also discuss the case when some devices are unavailable for inference. 

\subsubsection{\textbf{Homogeneous Devices}}
Table~\ref{table:homogeneous} presents the evaluation of NoNN-2S, NoNN-8S, and teacher models for both RPi's and Odroids. In this experiment, each student is deployed on a different device, so in the 8-student case we have eight separate devices (either RPi or Odroid). From the experiment in Section 3, since splitting the teacher model horizontally leads to more than $10\times$ increment in latency (even on powerful x86 machines), we executed the teacher on a single edge device, as otherwise it would take even longer. 
\begin{table*}[tb]
\centering
\caption{Accuracy, Performance, and Energy Results for CIFAR-10 on both Raspberry Pi and Odroid-XU4S}
\scalebox{0.75}{
\label{table:homogeneous}
\begin{tabular}{|l|c|c|c||c|c||c|c|c||c|c|}
\hline
\multirow{3}{*}{} & \multicolumn{5}{c||}{Raspberry Pi} & \multicolumn{5}{c|}{Odroid-XU4S} \\ \cline{2-11} 
 & \multirow{2}{*}{Teacher} & \multirow{2}{*}{2S} & \multirow{2}{*}{8S} & \multicolumn{2}{c||}{Improvement} & \multirow{2}{*}{Teacher} & \multirow{2}{*}{2S} & \multirow{2}{*}{8S} & \multicolumn{2}{c|}{Improvement} \\ \cline{5-6} \cline{10-11} 
 &  &  &  & 2S & 8S &  &  &  & 2S & 8S \\ \hline \hline
Accuracy & 95.49\% & 94.32\% & 95.02\% & -1.17\% & -0.47\% & 95.49\% & 94.32\% & 95.02\% & -1.17\% & -0.47\% \\ \hline
Latency (ms) & 1405 & 115 & 150 & 12.22$\times$ & 9.37$\times$ & 224 & 25 & 36 & 8.96$\times$ & 6.22$\times$ \\ \hline
Energy per node (mJ) & 3430.67$^{*}$ & 238.98 & 249.15 & 14.36$\times$ & 13.77$\times$ & 3084.08$^{*}$ & 219.07 & 237.45 & 14.08$\times$ & 12.99$\times$ \\ \hline \hline
Theoretical FLOPS per student & 2.6G$^{*}$ & 167M & 167M & 15.56$\times$ & 15.56$\times$  & 2.6G$^{*}$ & 167M & 167M & 15.56$\times$  & 15.56$\times$  \\ \hline
\end{tabular}
}
\begin{flushleft}
{\scriptsize $^{*}$Teacher model is executed on a single node since splitting the teacher model onto multiple devices incurs heavy communication latency (see Section 3). NoNN energy and FLOP numbers are reported per student because we are mostly concerned with per node budgets for memory and FLOPS. Of note, 2S (8S) refers to the NoNN-2S (NoNN-8S) model with two (eight) individual students deployed on two (eight) separate devices.}\\
\end{flushleft}
\end{table*}

On the RPi's, NoNN-2S improves the performance and energy per node by $12.22\times$ and $14.36\times$ w.r.t. teacher. Also, since NoNN-8S synchronizes eight devices, it achieves $9.37\times$ better latency and $13.77\times$ better energy per node. The same trend is observed for the Odroid boards, having $8.96\times$ and $14.08\times$ improvement for NoNN-2S in performance and energy per node, respectively, and $6.22\times$ and $12.99\times$ for NoNN-8S. Note that, the reduction in theoretical FLOPS for NoNN w.r.t. the teacher is about 15$\times$, while energy per node reduction for both RPi's and Odroids is 13-14$\times$. This shows very good agreement between theory and practice. Further, the latency and energy per node do \textit{not} increase significantly between NoNN-2S and NoNN-8S for both devices. Therefore, our proposed NoNN leads to significant gains in performance and per node energy, while maintaining accuracy within 0.5\%-1.17\% compared to the teacher.

Table~\ref{table:latency_breakdown} presents the average latency breakdown for computation and communication per image, considering the 2S and 8S cases. The computation time for one image increases by merely 4\% and 8\% for RPi and Odroid, respectively, when going from 2S to 8S, because each device operates in parallel. But when communication is considered, the overhead of sending and receiving data for 8S increases the average latency by $3.34\times$ for RPi's, and $2.57\times$ for Odroid boards (compared to the 2S-case). Again, this communication cost is very small compared to the cost incurred from directly splitting the single, large models. For instance, we have 43ms communication latency for distributing NoNN-8S on RPi's. In contrast, splitting the teacher on x86 incurs $1006-86=920$ms communication latency (see Section 3), which is $21\times$ higher than NoNN-8S. 

\begin{table}[tb]
\centering
\caption{Average Latency Breakdown Per Inference}
\scalebox{0.8}{
\label{table:latency_breakdown}
\begin{tabular}{|l|c|c||c|c||c|}
\hline
\multirow{2}{*}{Latency (ms)} & \multicolumn{2}{c||}{Raspberry Pi} & \multicolumn{2}{c||}{Odroid-XU4S} & x86 \\ \cline{2-6} 
 & 2S & 8S & 2S & 8S & 2-Split Teacher \\ \hline \hline
Computation & 101.91 & 106.24 & 19.01 & 20.60 & 86 \\ \hline
Communication & 13.10 & 43.76 & 6.00 & 15.40 & 920 \\ \hline
\end{tabular}
}
\end{table}

\subsubsection{\textbf{Comparison of Latency for Distributed Inference w.r.t. a Model Compression Baseline}}
Next, we present a concrete, side-by-side comparison between the distributed inference latency incurred by our proposed NoNN and ATKD~\cite{atkd}, a traditional model compression baseline. Note that, we compare the latency of NoNN against ATKD~\cite{atkd} as it achieves the best baseline accuracy results. Consequently, for this experiment, we use the ATKD model WRN40-2 (2.2M parameters, $95.03\%$ accuracy on CIFAR-10 dataset) distilled from the same teacher model as above (\textit{i.e.}, WRN40-4, 8.9M parameters, $95.49\%$ accuracy). To achieve higher accuracy, the size of the model compressed via ATKD grows significantly and, therefore, it must be distributed across multiple resource-constrained devices as it may not fit within the given per-device memory budget (henceforth known as the \textit{split-ATKD} experiment); similar to the split-teacher experiment above (see Section 3 and Table~\ref{table:latency_breakdown}), ATKD baseline incurs communication at each intermediate layer. Since TVM generates a single binary file for the entire deep network, models compiled in TVM cannot be used to perform such split-teacher or split-ATKD experiments on RPi's (as we need to extract the intermediate outputs at each layer and communicate them across devices). Hence, for the following split-ATKD experiments, we install Pytorch 0.4.1 on all eight RPi's to run distributed inference for both split-ATKD and NoNN-8S.

Table~\ref{table:splitATKD} shows the accuracy, parameters per device, FLOPS executed per device, and total latency for NoNN-8S and split-ATKD models in pytorch on RPi's. We distribute the ATKD baseline in two different ways:
\begin{table}[tb]
\centering
\caption{NoNN and Split-ATKD Latency on RPi's (Pytorch)}
\scalebox{0.8}{
\label{table:splitATKD}
\begin{tabular}{|l|c|c||c|}
\hline
\multirow{2}{*}{} & \multicolumn{2}{c||}{Split-ATKD (WRN40-2)} & NoNN-8S \\ \cline{2-3} 
 & 4 RPi's & 8 RPi's & 8 RPi's\\ \hline \hline
Accuracy & $95.03\%$ & $95.03\%$ & $95.02\%$ \\ \hline
Parameters per device & $550$K & $275$K & $430$K \\ \hline
FLOPS per device & $163$M & $82$M & $167$M \\ \hline
Total latency per inference (s) & $23$ & $28.5$ & $\bm{0.85}$ \\ \hline \hline
Speedup with NoNN & $\bm{27\times}$ & $\bm{33\times}$ & $-$ \\ \hline
\end{tabular}
}
\end{table}
\begin{enumerate}
    \item \textbf{ATKD split on four RPi's (split-ATKD-4S):} Each RPi for this case stores about 550K (2.2M/4) parameters and executes about 163M (655M/4) FLOPS. This experiment simulates the scenario where each device has, say, close to 500K-parameter memory-budget (similar to the 430K parameter-budget considered in NoNN for the CIFAR-10 dataset).
    \item \textbf{ATKD split on eight RPi's (split-ATKD-8S):} Each RPi for this case stores around 275K (2.2M/8) parameters and executes merely 82M (655M/8) FLOPS. Similar to NoNN-8S, this case compares the performance of split-ATKD on eight separate devices.
\end{enumerate}
Note that, the split-ATKD-4S case has slightly more parameters than the single NoNN-8S student (550K \textit{vs.} 430K) but executes slightly fewer FLOPS than NoNN-8S (163M \textit{vs.} 167M). On the other hand, the split-ATKD-8S case executes far fewer FLOPS than a single NoNN-8S student (82M \textit{vs.} 167M) and also stores fewer parameters than NoNN-8S (275K vs. 430K). As evident from Table~\ref{table:splitATKD}, despite the lower computation involved in both the split-ATKD cases, the heavy communication at each layer incurred by split-ATKD-4S (split-ATKD-8S) results in a total latency of $23$s ($28.5$s). In contrast, NoNN-8S takes only $0.85$s (in pytorch) to perform the distributed inference\footnote{The total latency for distributed inference via NoNN-8S increases from $150$ms in TVM (see Table~\ref{table:homogeneous}) to $850$ms in pytorch. This shows that the device-level optimizations performed by TVM are important for deep learning inference on edge devices.}, which is $27\times$ ($33\times$) better than the split-ATKD-4S (split-ATKD-8S) baseline, while achieving nearly the same accuracy. Hence, this result emphasizes the importance of our proposed NoNN for distributed deep learning inference on IoT. 

\subsubsection{\textbf{Heterogeneous Devices}}
To demonstrate NoNN's flexibility, we perform the following experiment: we start with a homogeneous scenario for the RPi's, then, gradually move each student to an Odroid board, depicting an environment where we have several small devices (RPi's) and one slightly more powerful device (Odroid), the latter running on the big cores to maximize performance. Each small device (RPi) executes only one student, while the more powerful device (Odroid) executes multiple students.

\begin{figure*}[tb]
    \begin{subfigure}{0.44\textwidth}
    \centering
    \includegraphics[width=1.0\textwidth]{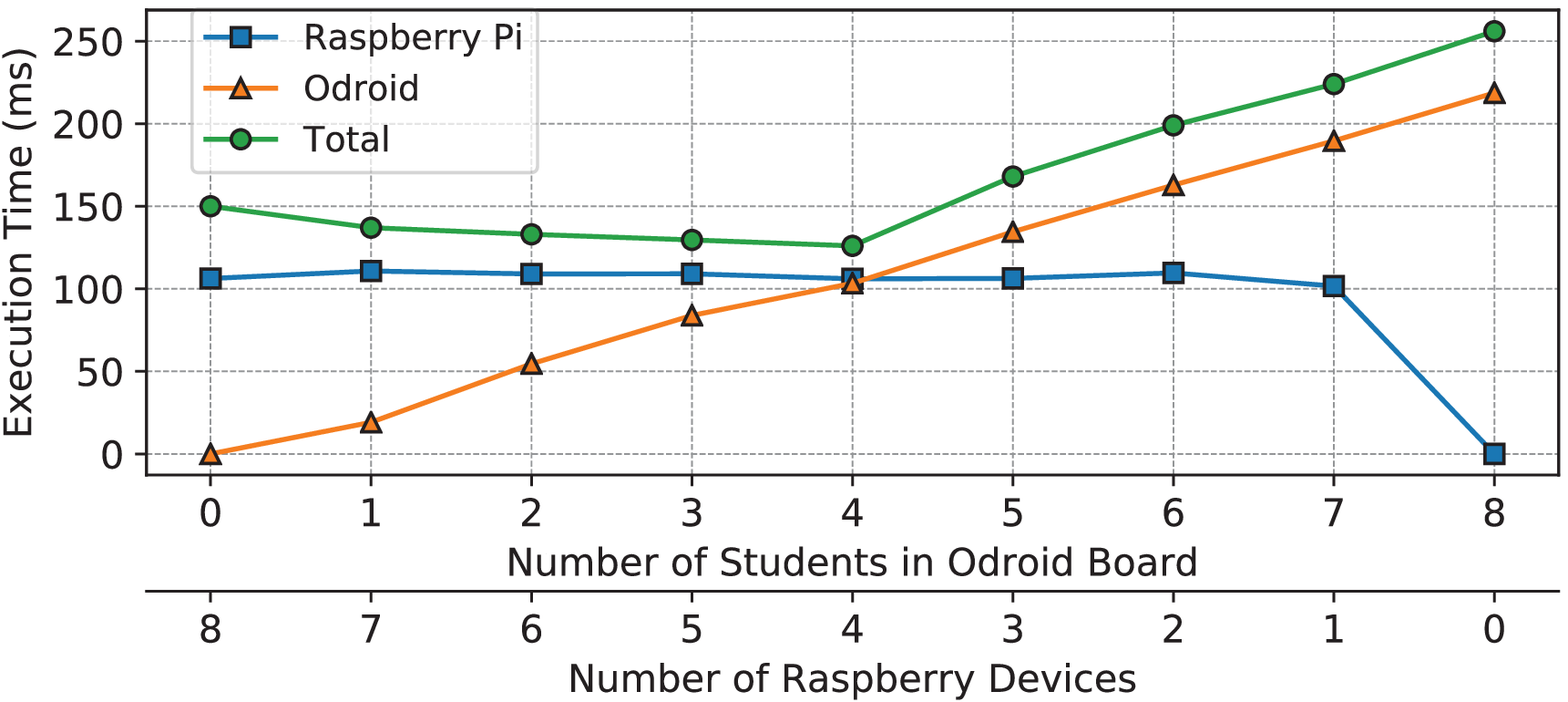}
    \caption{Performance per Inference}
    \label{fig:latency}
    \end{subfigure}%
    \hspace{5mm}
    \begin{subfigure}{0.44\textwidth}
    \centering
    \includegraphics[width=1.0\textwidth]{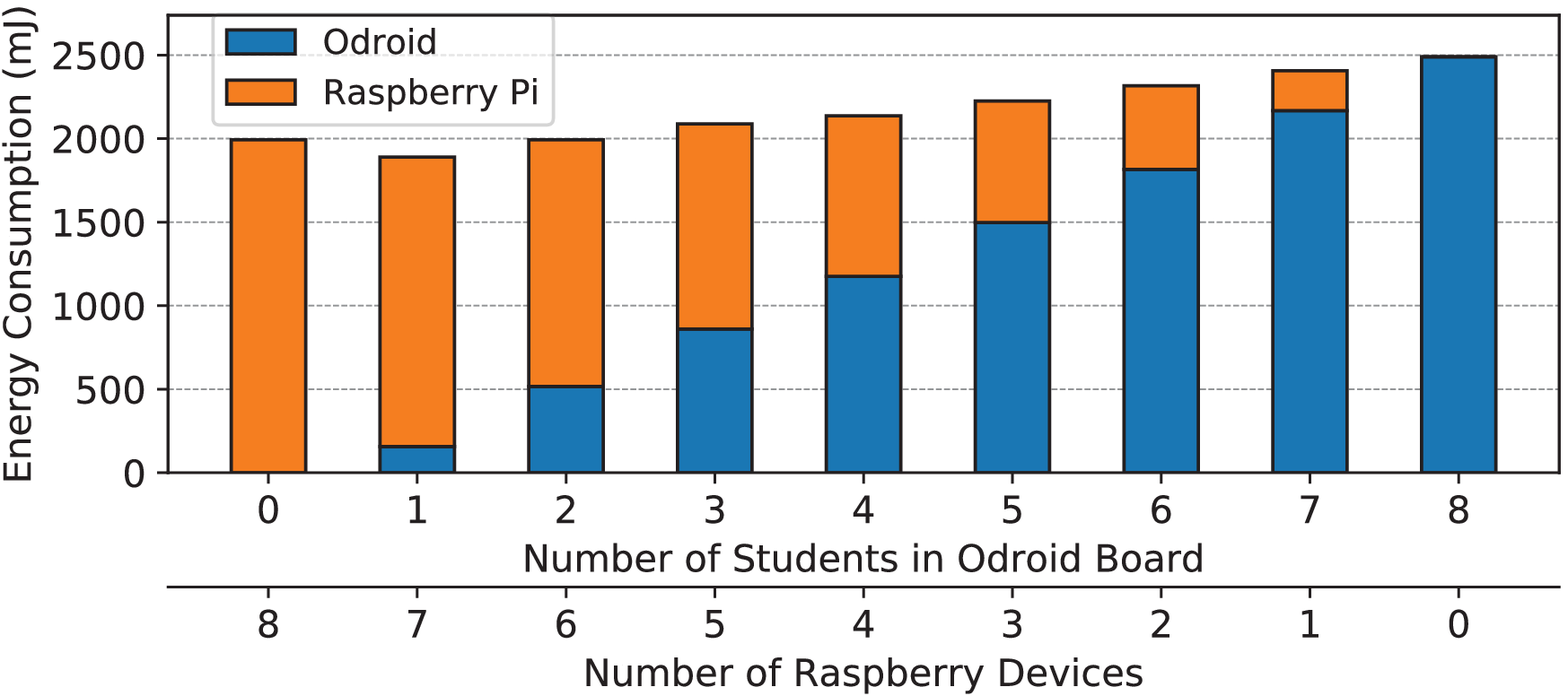}
    \caption{Energy Consumption per Inference}
    \label{fig:energy}
    \end{subfigure}%
    \caption{Performance and energy as the number of Raspberry devices is reduced and the workload for the Odroid board is increased, always executing a total of eight students.}
    \label{fig:latency_and_energy}
\end{figure*}

Fig.~\ref{fig:latency} breaks down the latency into three parts: latency of RPi's, the single Odroid, and the \textit{total latency} including the server. The total latency represents the execution time, starting when the server transmits the first image and ending when fc layer is applied (after receiving and concatenating results from all students). The RPi performance represents the latency (\textit{i.e.}, both computation and communication time) for the \textit{slowest} device. 

As the Odroids are considerably faster than the RPi's in the current setup, a single Odroid is able to execute four students in the same amount of time a RPi takes to compute one student (Fig.~\ref{fig:latency}). As the number of students in the Odroid increases even further, its latency becomes a bottleneck and the total latency increases. The difference between the total latency and the slowest device is the communication time between the server and the devices, which is 30.65ms on average. The server (x86 machine) takes 0.1ms to concatenate and apply the fc layer, which is negligible. Note that, until four students are deployed in Odroid, the total latency in Fig.~\ref{fig:latency} keeps on reducing since the server can process the Odroid results while the RPi's are still running the individual students. Hence, performing this task in parallel reduces the critical path.

Fig.~\ref{fig:energy} depicts the energy consumption for \textit{all} RPi devices and the single Odroid device. As evident, we have a trade-off among performance, energy, and number of devices. For instance, when the Odroid runs just one student, the total energy gets minimized because Odroid is more energy-efficient than the RPi for the one student case. However, as we increase the workload for the Odroid (\textit{i.e.}, number of students), the total energy increases even though the number of devices (RPi's) decreases. In the current setup, inference on eight separate RPi's consumes the same amount of energy as using six RPi's plus two students on the Odroid. Comparing the two extreme cases (eight students in eight RPi's, or all eight students in a single Odroid), the Odroid consumes 25\% more energy. 

Therefore, this evaluation demonstrates three important aspects: (\textit{i}) NoNN approach demonstrates good performance in both homogeneous and heterogeneous environments,  (\textit{ii}) In terms of energy, distributing the students on several devices instead of using a single more powerful device can reduce the total energy consumption for inference. (\textit{iii}) Finally, to maximize the energy savings for the entire system, deploying NoNN on a combination of several devices can trade-off between energy dissipation and number of devices. 

\subsubsection{\textbf{Robustness of NoNN}}
We now evaluate a scenario where some devices may become unavailable due to a variety of reasons such as power depletion or processor/network failure. We evaluate all possible cases in which such devices may fail. For instance, 
for six active devices, any two devices may fail, resulting in 28 possible scenarios, and so on for the remaining cases.

\begin{figure}[tb]
    \centering
    \includegraphics[width=0.6\textwidth]{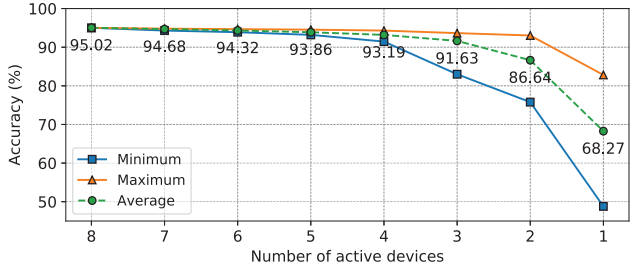}
    \caption{Accuracy as some devices become unavailable due to device failures (\textit{e.g.}, due to processor or network failures, and battery depletion). NoNN achieves more than $91\%$ accuracy, on average, when only three devices are available. When four devices are available, NoNN achieves more than $93\%$ accuracy.}
    \label{fig:accuracy}
\end{figure}

Fig.~\ref{fig:accuracy} shows the accuracy drop as we remove some devices. On an average, NoNN is still able to achieve more than 91\% accuracy when only three devices are available. Moreover, if a minimum of four devices are available, the average accuracy is more than 93\%. As more devices become unavailable, the difference between the minimum and the maximum accuracy increases since the final decision depends on output from all devices. When only a few devices are available, the final result depends on the characteristics of each student, some being able to deliver higher accuracy than others. For instance, one of the students is able to achieve 82.82\% accuracy alone (\textit{i.e.}, the maximum point when only one student is active in Fig.~\ref{fig:accuracy}). Therefore, deploying this student on the primary device (which will always be available) will guarantee at least 82.82\% accuracy. Other devices can then be used to further improve the accuracy of the system. Hence, NoNN is robust to random device failures. Of note, robustness can also be formulated as part of the partitioning problem itself; this, however, is left as a future work.

\section{Conclusion and Future Work}
In this paper, we have proposed a new NoNN paradigm to compress a large teacher model into multiple highly-compressed student modules that can be distributed across a network of edge devices with minimal communication overhead. To this end, we have first proposed a network science-based partitioning of teacher's knowledge, and then trained individual students on the corresponding partitions. 

With extensive experiments, we have demonstrated that NoNN achieves close to teacher's accuracy with significantly lower memory ($2.5\times$-$24\times$ gain w.r.t. teacher) and computation ($2\times$-$15\times$ fewer FLOPS w.r.t. teacher), while guaranteeing that individual modules of NoNN fit within some given memory/FLOP budget. We have also shown that NoNN achieves higher accuracy than several baselines with no communication among students until the last layer. 

Finally, we have deployed the proposed models for CIFAR-10 dataset on Raspberry Pi and Odroid, and have demonstrated $6.22\times$-$12.22\times$ improvement in performance and $12.99\times$-$14.36\times$ in energy per node w.r.t. teacher. We have further shown that NoNN model results in up to $33\times$ reduction in total latency for distributed inference on multiple edge devices w.r.t. a state-of-the-art model compression baseline. Hence, the proposed communication-aware model compression can ultimately lead to effective deployment of deep networks on multiple memory-constrained IoT-devices.

The current work opens up several new directions. In future, we plan to address problems like how to accurately map deep network partitions to heterogeneous devices, and also make robustness as a primary objective of NoNN. We further plan to explore NoNN for other types of deep learning models such as RNNs.

%
%
%

\begin{acks}
This material is based on research sponsored in part by Air Force Research Laboratory (AFRL) and Defense Advanced Research Projects Agency (DARPA) under agreement number FA8650-18-2-7860. The U.S. Government is authorized to reproduce and distribute reprints for Governmental purposes notwithstanding any copyright notation thereon. The views and conclusions contained herein are those of the authors and should not be interpreted as necessarily representing the official policies or endorsements, either expressed or implied, of Air Force Research Laboratory (AFRL) and Defense Advanced Research Projects Agency (DARPA) or the U.S. Government.

We thank the anonymous reviewers for their useful suggestions which helped improve the paper. We further thank J. Wang, I. Pardesi, and W. Chen of Carnegie Mellon University for their help with the experimental part. We also thank NVIDIA Corporation for donating a Titan Xp GPU which was used for conducting some experiments. Finally, we acknowledge Amazon support through the AWS ML Research Program, and FastAI (\url{https://www.fast.ai/}) for releasing code to accelerate training on Imagenet dataset using AWS machines.
\end{acks}

\bibliographystyle{ACM-Reference-Format}
\bibliography{sample-bibliography}


\begin{thebibliography}{28}


\ifx \showCODEN    \undefined \def \showCODEN     #1{\unskip}     \fi
\ifx \showDOI      \undefined \def \showDOI       #1{#1}\fi
\ifx \showISBNx    \undefined \def \showISBNx     #1{\unskip}     \fi
\ifx \showISBNxiii \undefined \def \showISBNxiii  #1{\unskip}     \fi
\ifx \showISSN     \undefined \def \showISSN      #1{\unskip}     \fi
\ifx \showLCCN     \undefined \def \showLCCN      #1{\unskip}     \fi
\ifx \shownote     \undefined \def \shownote      #1{#1}          \fi
\ifx \showarticletitle \undefined \def \showarticletitle #1{#1}   \fi
\ifx \showURL      \undefined \def \showURL       {\relax}        \fi
\providecommand\bibfield[2]{#2}
\providecommand\bibinfo[2]{#2}
\providecommand\natexlab[1]{#1}
\providecommand\showeprint[2][]{arXiv:#2}

\bibitem[\protect\citeauthoryear{Ba and Caruana}{Ba and Caruana}{2014}]%
        {deepShallow1}
\bibfield{author}{\bibinfo{person}{Jimmy Ba} {and} \bibinfo{person}{Rich
  Caruana}.} \bibinfo{year}{2014}\natexlab{}.
\newblock \showarticletitle{Do deep nets really need to be deep?}. In
  \bibinfo{booktitle}{\emph{Advances in neural information processing
  systems}}. \bibinfo{pages}{2654--2662}.
\newblock


\bibitem[\protect\citeauthoryear{Facebook}{Facebook}{2017}]%
        {ONNX}
\bibfield{author}{\bibinfo{person}{Facebook}.} \bibinfo{year}{2017}\natexlab{}.
\newblock \bibinfo{title}{{ONNX: Open Neural Network Exchange Format}}.
\newblock
\newblock
\urldef\tempurl%
\url{https://onnx.ai/}
\showURL{%
\tempurl}


\bibitem[\protect\citeauthoryear{Gao, Wang, and Ji}{Gao et~al\mbox{.}}{2018}]%
        {channelnets}
\bibfield{author}{\bibinfo{person}{Hongyang Gao}, \bibinfo{person}{Zhengyang
  Wang}, {and} \bibinfo{person}{Shuiwang Ji}.} \bibinfo{year}{2018}\natexlab{}.
\newblock \showarticletitle{ChannelNets: Compact and Efficient Convolutional
  Neural Networks via Channel-Wise Convolutions}. In
  \bibinfo{booktitle}{\emph{Advances in Neural Information Processing
  Systems}}. \bibinfo{pages}{5203--5211}.
\newblock


\bibitem[\protect\citeauthoryear{Han, Mao, and Dally}{Han
  et~al\mbox{.}}{2015a}]%
        {deepComp}
\bibfield{author}{\bibinfo{person}{Song Han}, \bibinfo{person}{Huizi Mao},
  {and} \bibinfo{person}{William~J Dally}.} \bibinfo{year}{2015}\natexlab{a}.
\newblock \showarticletitle{Deep compression: Compressing deep neural networks
  with pruning, trained quantization and Huffman coding}.
\newblock \bibinfo{journal}{\emph{arXiv:1510.00149}} (\bibinfo{year}{2015}).
\newblock


\bibitem[\protect\citeauthoryear{Han, Pool, Tran, and Dally}{Han
  et~al\mbox{.}}{2015b}]%
        {hanNIPS2015}
\bibfield{author}{\bibinfo{person}{Song Han}, \bibinfo{person}{Jeff Pool},
  \bibinfo{person}{John Tran}, {and} \bibinfo{person}{William Dally}.}
  \bibinfo{year}{2015}\natexlab{b}.
\newblock \showarticletitle{Learning both weights and connections for efficient
  neural network}. In \bibinfo{booktitle}{\emph{NIPS}}.
  \bibinfo{pages}{1135--1143}.
\newblock


\bibitem[\protect\citeauthoryear{Hinton, Vinyals, and Dean}{Hinton
  et~al\mbox{.}}{2015}]%
        {hintonKD}
\bibfield{author}{\bibinfo{person}{Geoffrey Hinton}, \bibinfo{person}{Oriol
  Vinyals}, {and} \bibinfo{person}{Jeff Dean}.}
  \bibinfo{year}{2015}\natexlab{}.
\newblock \showarticletitle{Distilling the knowledge in a neural network}.
\newblock \bibinfo{journal}{\emph{arXiv:1503.02531}} (\bibinfo{year}{2015}).
\newblock


\bibitem[\protect\citeauthoryear{Howard}{Howard}{2018}]%
        {fastai}
\bibfield{author}{\bibinfo{person}{Jeremy Howard}.}
  \bibinfo{year}{2018}\natexlab{}.
\newblock \showarticletitle{Imagenet in 18 minutes}.
\newblock
  \bibinfo{howpublished}{\url{https://www.fast.ai/2018/08/10/fastai-diu-imagenet/}}.
\newblock  (\bibinfo{year}{2018}).
\newblock
\newblock
\shownote{Accessed: 2018-10-01.}


\bibitem[\protect\citeauthoryear{Hubara and et~al.}{Hubara and et~al.}{2017}]%
        {quant2}
\bibfield{author}{\bibinfo{person}{Itay Hubara} {and} \bibinfo{person}{et al.}}
  \bibinfo{year}{2017}\natexlab{}.
\newblock \showarticletitle{Quantized neural networks: Training neural networks
  with low precision weights and activations}.
\newblock \bibinfo{journal}{\emph{JMLR}} \bibinfo{volume}{18},
  \bibinfo{number}{1} (\bibinfo{year}{2017}), \bibinfo{pages}{6869--6898}.
\newblock


\bibitem[\protect\citeauthoryear{Iandola, Han, Moskewicz, Ashraf, Dally, and
  Keutzer}{Iandola et~al\mbox{.}}{2016}]%
        {sqn}
\bibfield{author}{\bibinfo{person}{Forrest~N Iandola}, \bibinfo{person}{Song
  Han}, \bibinfo{person}{Matthew~W Moskewicz}, \bibinfo{person}{Khalid Ashraf},
  \bibinfo{person}{William~J Dally}, {and} \bibinfo{person}{Kurt Keutzer}.}
  \bibinfo{year}{2016}\natexlab{}.
\newblock \showarticletitle{SqueezeNet: AlexNet-level accuracy with 50x fewer
  parameters and< 0.5 MB model size}.
\newblock \bibinfo{journal}{\emph{arXiv:1602.07360}} (\bibinfo{year}{2016}).
\newblock


\bibitem[\protect\citeauthoryear{Kim, Park, Kim, and Hwang}{Kim
  et~al\mbox{.}}{2017}]%
        {splitnet}
\bibfield{author}{\bibinfo{person}{Juyong Kim}, \bibinfo{person}{Yookoon Park},
  \bibinfo{person}{Gunhee Kim}, {and} \bibinfo{person}{Sung~Ju Hwang}.}
  \bibinfo{year}{2017}\natexlab{}.
\newblock \showarticletitle{SplitNet: Learning to semantically split deep
  networks for parameter reduction and model parallelization}. In
  \bibinfo{booktitle}{\emph{International Conference on Machine Learning}}.
  \bibinfo{pages}{1866--1874}.
\newblock


\bibitem[\protect\citeauthoryear{Kim and Rush}{Kim and Rush}{2016}]%
        {seqKD}
\bibfield{author}{\bibinfo{person}{Yoon Kim} {and} \bibinfo{person}{Alexander~M
  Rush}.} \bibinfo{year}{2016}\natexlab{}.
\newblock \showarticletitle{Sequence-level knowledge distillation}.
\newblock \bibinfo{journal}{\emph{arXiv preprint arXiv:1606.07947}}
  (\bibinfo{year}{2016}).
\newblock


\bibitem[\protect\citeauthoryear{Lai, Suda, and Chandra}{Lai
  et~al\mbox{.}}{2017}]%
        {floatFix}
\bibfield{author}{\bibinfo{person}{Liangzhen Lai}, \bibinfo{person}{Naveen
  Suda}, {and} \bibinfo{person}{Vikas Chandra}.}
  \bibinfo{year}{2017}\natexlab{}.
\newblock \showarticletitle{Deep Convolutional Neural Network Inference with
  Floating-point Weights and Fixed-point Activations}.
\newblock \bibinfo{journal}{\emph{arXiv:1703.03073}} (\bibinfo{year}{2017}).
\newblock


\bibitem[\protect\citeauthoryear{Li, Kadav, Durdanovic, Samet, and Graf}{Li
  et~al\mbox{.}}{2016}]%
        {prune2}
\bibfield{author}{\bibinfo{person}{Hao Li}, \bibinfo{person}{Asim Kadav},
  \bibinfo{person}{Igor Durdanovic}, \bibinfo{person}{Hanan Samet}, {and}
  \bibinfo{person}{Hans~Peter Graf}.} \bibinfo{year}{2016}\natexlab{}.
\newblock \showarticletitle{Pruning filters for efficient convnets}.
\newblock \bibinfo{journal}{\emph{arXiv:1608.08710}} (\bibinfo{year}{2016}).
\newblock


\bibitem[\protect\citeauthoryear{Mao and et~al.}{Mao and et~al.}{2017}]%
        {modnn}
\bibfield{author}{\bibinfo{person}{Jiachen Mao} {and} \bibinfo{person}{et al.}}
  \bibinfo{year}{2017}\natexlab{}.
\newblock \showarticletitle{Modnn: Local distributed mobile computing system
  for deep neural network}. In \bibinfo{booktitle}{\emph{2017 DATE
  Conference}}. IEEE, \bibinfo{pages}{1396--1401}.
\newblock


\bibitem[\protect\citeauthoryear{Mao, Yang, Wen, Wu, Song, Nixon, Chen, Li, and
  Chen}{Mao et~al\mbox{.}}{2017}]%
        {mednn}
\bibfield{author}{\bibinfo{person}{Jiachen Mao}, \bibinfo{person}{Zhongda
  Yang}, \bibinfo{person}{Wei Wen}, \bibinfo{person}{Chunpeng Wu},
  \bibinfo{person}{Linghao Song}, \bibinfo{person}{Kent~W Nixon},
  \bibinfo{person}{Xiang Chen}, \bibinfo{person}{Hai Li}, {and}
  \bibinfo{person}{Yiran Chen}.} \bibinfo{year}{2017}\natexlab{}.
\newblock \showarticletitle{Mednn: A distributed mobile system with enhanced
  partition and deployment for large-scale dnns}. In
  \bibinfo{booktitle}{\emph{Proceedings of the 36th International Conference on
  Computer-Aided Design}}. IEEE Press, \bibinfo{pages}{751--756}.
\newblock


\bibitem[\protect\citeauthoryear{Newman, Barabasi, and Watts}{Newman
  et~al\mbox{.}}{2011}]%
        {networksci}
\bibfield{author}{\bibinfo{person}{Mark Newman}, \bibinfo{person}{Albert-Laszlo
  Barabasi}, {and} \bibinfo{person}{Duncan~J Watts}.}
  \bibinfo{year}{2011}\natexlab{}.
\newblock \bibinfo{booktitle}{\emph{The structure and dynamics of networks}}.
  Vol.~\bibinfo{volume}{19}.
\newblock \bibinfo{publisher}{Princeton University Press}.
\newblock


\bibitem[\protect\citeauthoryear{Newman}{Newman}{2006}]%
        {newmanComm}
\bibfield{author}{\bibinfo{person}{Mark~EJ Newman}.}
  \bibinfo{year}{2006}\natexlab{}.
\newblock \showarticletitle{Modularity and community structure in networks}.
\newblock \bibinfo{journal}{\emph{Proceedings of the national academy of
  sciences}} \bibinfo{volume}{103}, \bibinfo{number}{23}
  (\bibinfo{year}{2006}), \bibinfo{pages}{8577--8582}.
\newblock


\bibitem[\protect\citeauthoryear{Quattoni and Torralba}{Quattoni and
  Torralba}{2009}]%
        {scen}
\bibfield{author}{\bibinfo{person}{Ariadna Quattoni} {and}
  \bibinfo{person}{Antonio Torralba}.} \bibinfo{year}{2009}\natexlab{}.
\newblock \showarticletitle{Recognizing indoor scenes}. In
  \bibinfo{booktitle}{\emph{2009 IEEE Conference on Computer Vision and Pattern
  Recognition}}. IEEE, \bibinfo{pages}{413--420}.
\newblock


\bibitem[\protect\citeauthoryear{Sandler and et~al.}{Sandler and
  et~al.}{2018}]%
        {mobilenetV2}
\bibfield{author}{\bibinfo{person}{Mark Sandler} {and} \bibinfo{person}{et
  al.}} \bibinfo{year}{2018}\natexlab{}.
\newblock \showarticletitle{Inverted residuals and linear bottlenecks: Mobile
  networks for classification, detection and segmentation}.
\newblock \bibinfo{journal}{\emph{arXiv:1801.04381}} (\bibinfo{year}{2018}).
\newblock


\bibitem[\protect\citeauthoryear{STMicro}{STMicro}{2018}]%
        {stMicro}
\bibfield{author}{\bibinfo{person}{STMicro}.} \bibinfo{year}{2018}\natexlab{}.
\newblock \bibinfo{title}{{Datasheet for Arm-Based Microcontroller with up to
  512KB total storage (including FLASH memory). Product Page:
  \url{https://bit.ly/2I5ZSMR}. Datasheet}}.
\newblock
\newblock
\urldef\tempurl%
\url{https://bit.ly/2Kz8ehD}
\showURL{%
\tempurl}


\bibitem[\protect\citeauthoryear{Tang, Wang, and Zhang}{Tang
  et~al\mbox{.}}{2016}]%
        {rnnKD}
\bibfield{author}{\bibinfo{person}{Zhiyuan Tang}, \bibinfo{person}{Dong Wang},
  {and} \bibinfo{person}{Zhiyong Zhang}.} \bibinfo{year}{2016}\natexlab{}.
\newblock \showarticletitle{Recurrent neural network training with dark
  knowledge transfer}. In \bibinfo{booktitle}{\emph{2016 IEEE International
  Conference on Acoustics, Speech and Signal Processing (ICASSP)}}. IEEE,
  \bibinfo{pages}{5900--5904}.
\newblock


\bibitem[\protect\citeauthoryear{Welinder, Branson, Mita, Wah, Schroff,
  Belongie, and Perona}{Welinder et~al\mbox{.}}{2010}]%
        {cub}
\bibfield{author}{\bibinfo{person}{Peter Welinder}, \bibinfo{person}{Steve
  Branson}, \bibinfo{person}{Takeshi Mita}, \bibinfo{person}{Catherine Wah},
  \bibinfo{person}{Florian Schroff}, \bibinfo{person}{Serge Belongie}, {and}
  \bibinfo{person}{Pietro Perona}.} \bibinfo{year}{2010}\natexlab{}.
\newblock \showarticletitle{Caltech-UCSD birds 200}.
\newblock  (\bibinfo{year}{2010}).
\newblock


\bibitem[\protect\citeauthoryear{Yang and et~al.}{Yang and et~al.}{2016}]%
        {prune3}
\bibfield{author}{\bibinfo{person}{Tien-Ju Yang} {and} \bibinfo{person}{et
  al.}} \bibinfo{year}{2016}\natexlab{}.
\newblock \showarticletitle{Designing energy-efficient convolutional neural
  networks using energy-aware pruning}.
\newblock \bibinfo{journal}{\emph{arXiv:1611.05128}} (\bibinfo{year}{2016}).
\newblock


\bibitem[\protect\citeauthoryear{Zagoruyko and Komodakis}{Zagoruyko and
  Komodakis}{2016}]%
        {wrn}
\bibfield{author}{\bibinfo{person}{Sergey Zagoruyko} {and}
  \bibinfo{person}{Nikos Komodakis}.} \bibinfo{year}{2016}\natexlab{}.
\newblock \showarticletitle{Wide residual networks}.
\newblock \bibinfo{journal}{\emph{BMVC}} (\bibinfo{year}{2016}).
\newblock


\bibitem[\protect\citeauthoryear{Zagoruyko and Komodakis}{Zagoruyko and
  Komodakis}{2017}]%
        {atkd}
\bibfield{author}{\bibinfo{person}{Sergey Zagoruyko} {and}
  \bibinfo{person}{Nikos Komodakis}.} \bibinfo{year}{2017}\natexlab{}.
\newblock \showarticletitle{Improving the performance of convolutional neural
  networks via attention transfer}.
\newblock \bibinfo{journal}{\emph{ICLR}} (\bibinfo{year}{2017}).
\newblock


\bibitem[\protect\citeauthoryear{Zhang and et~al.}{Zhang and et~al.}{2017}]%
        {shufflenets}
\bibfield{author}{\bibinfo{person}{Xiangyu Zhang} {and} \bibinfo{person}{et
  al.}} \bibinfo{year}{2017}\natexlab{}.
\newblock \showarticletitle{ShuffleNet: An Extremely Efficient Convolutional
  Neural Network for Mobile Devices.}
\newblock \bibinfo{journal}{\emph{CoRR}}  \bibinfo{volume}{abs/1707.01083}
  (\bibinfo{year}{2017}).
\newblock


\bibitem[\protect\citeauthoryear{Zhang, Suda, Lai, and Chandra}{Zhang
  et~al\mbox{.}}{2017}]%
        {kws}
\bibfield{author}{\bibinfo{person}{Yundong Zhang}, \bibinfo{person}{Naveen
  Suda}, \bibinfo{person}{Liangzhen Lai}, {and} \bibinfo{person}{Vikas
  Chandra}.} \bibinfo{year}{2017}\natexlab{}.
\newblock \showarticletitle{Hello Edge: Keyword Spotting on Microcontrollers}.
\newblock \bibinfo{journal}{\emph{arXiv:1711.07128}} (\bibinfo{year}{2017}).
\newblock


\bibitem[\protect\citeauthoryear{Zhao, Barijough, and Gerstlauer}{Zhao
  et~al\mbox{.}}{2018}]%
        {deepthings}
\bibfield{author}{\bibinfo{person}{Zhuoran Zhao},
  \bibinfo{person}{Kamyar~Mirzazad Barijough}, {and} \bibinfo{person}{Andreas
  Gerstlauer}.} \bibinfo{year}{2018}\natexlab{}.
\newblock \showarticletitle{DeepThings: Distributed Adaptive Deep Learning
  Inference on Resource-Constrained IoT Edge Clusters}.
\newblock \bibinfo{journal}{\emph{IEEE Transactions on Computer-Aided Design of
  Integrated Circuits and Systems}} \bibinfo{volume}{37}, \bibinfo{number}{11}
  (\bibinfo{year}{2018}), \bibinfo{pages}{2348--2359}.
\newblock


\end{thebibliography}

%
%
%
%
%
%
%
%

\end{document}